\documentclass[lettersize,journal]{IEEEtran}
\usepackage{amsmath,amsfonts}
\usepackage{algorithm}
\usepackage{array}
\usepackage[caption=false,font=normalsize,labelfont=sf,textfont=sf]{subfig}
\usepackage{textcomp}
\usepackage{stfloats}
\usepackage{url}
\usepackage{verbatim}
\usepackage{graphicx}
\usepackage{cite}
\usepackage{amsmath}
\usepackage{amssymb}
\usepackage{booktabs}
\usepackage{times}
\usepackage{soul}
\usepackage{inputenc}
\usepackage{amsmath}
\usepackage{amsthm}
\usepackage{amsfonts}
\usepackage{booktabs}
\usepackage[switch]{lineno}
\usepackage{color}
\usepackage{multirow}
\usepackage{float}
\usepackage{colortbl}
\usepackage[colorlinks,linkcolor=red,anchorcolor=black,citecolor=green]{hyperref}
\usepackage{algorithmicx}
\usepackage{algpseudocode}
\usepackage[table,xcdraw]{xcolor}
\begin{document}

\title{DiffCR: A Fast Conditional Diffusion Framework for Cloud Removal from Optical Satellite Images}

\author{Xuechao Zou, Kai Li,~\IEEEmembership{Student Member, IEEE}, Junliang Xing,~\IEEEmembership{Senior Member, IEEE}, Yu Zhang, Shiying Wang, Lei Jin, and Pin Tao,~\IEEEmembership{Member, IEEE}
\thanks{$^*$Xuechao Zou and Kai Li have contributed equally to this work.}
\thanks{This paper was produced by the IEEE Publication Technology Group. They are in Piscataway, NJ.}
\thanks{Manuscript received April 19, 2021; revised August 16, 2021.}}

\markboth{IEEE Transactions on Geoscience and Remote Sensing,~Vol.~14, No.~8, August~2021}%
{Shell \MakeLowercase{\textit{et al.}}: A Sample Article Using IEEEtran.cls for IEEE Journals}


\maketitle

\begin{abstract}
Optical satellite images are a critical data source; however, cloud cover often compromises their quality, hindering image applications and analysis. Consequently, effectively removing clouds from optical satellite images has emerged as a prominent research direction. While recent advancements in cloud removal primarily rely on generative adversarial networks, which may yield suboptimal image quality, diffusion models have demonstrated remarkable success in diverse image-generation tasks, showcasing their potential in addressing this challenge. This paper presents a novel framework called DiffCR, which leverages conditional guided diffusion with deep convolutional networks for high-performance cloud removal for optical satellite imagery. Specifically, we introduce a decoupled encoder for conditional image feature extraction, providing a robust color representation to ensure the close similarity of appearance information between the conditional input and the synthesized output. Moreover, we propose a novel and efficient time and condition fusion block within the cloud removal model to accurately simulate the correspondence between the appearance in the conditional image and the target image at a low computational cost. Extensive experimental evaluations on two commonly used benchmark datasets demonstrate that DiffCR consistently achieves state-of-the-art performance on all metrics, with parameter and computational complexities amounting to only 5.1\% and 5.4\%, respectively, of those previous best methods. The source code, pre-trained models, and all the experimental results will be publicly available at \url{https://github.com/XavierJiezou/DiffCR} upon the paper's acceptance of this work. 
\end{abstract}

\begin{IEEEkeywords}
Cloud removal, diffusion models, optical satellite images, deep learning, autoencoder.
\end{IEEEkeywords}

\section{Introduction}\label{intro}


\begin{figure}[!t]
\centering
\includegraphics[width=\linewidth]{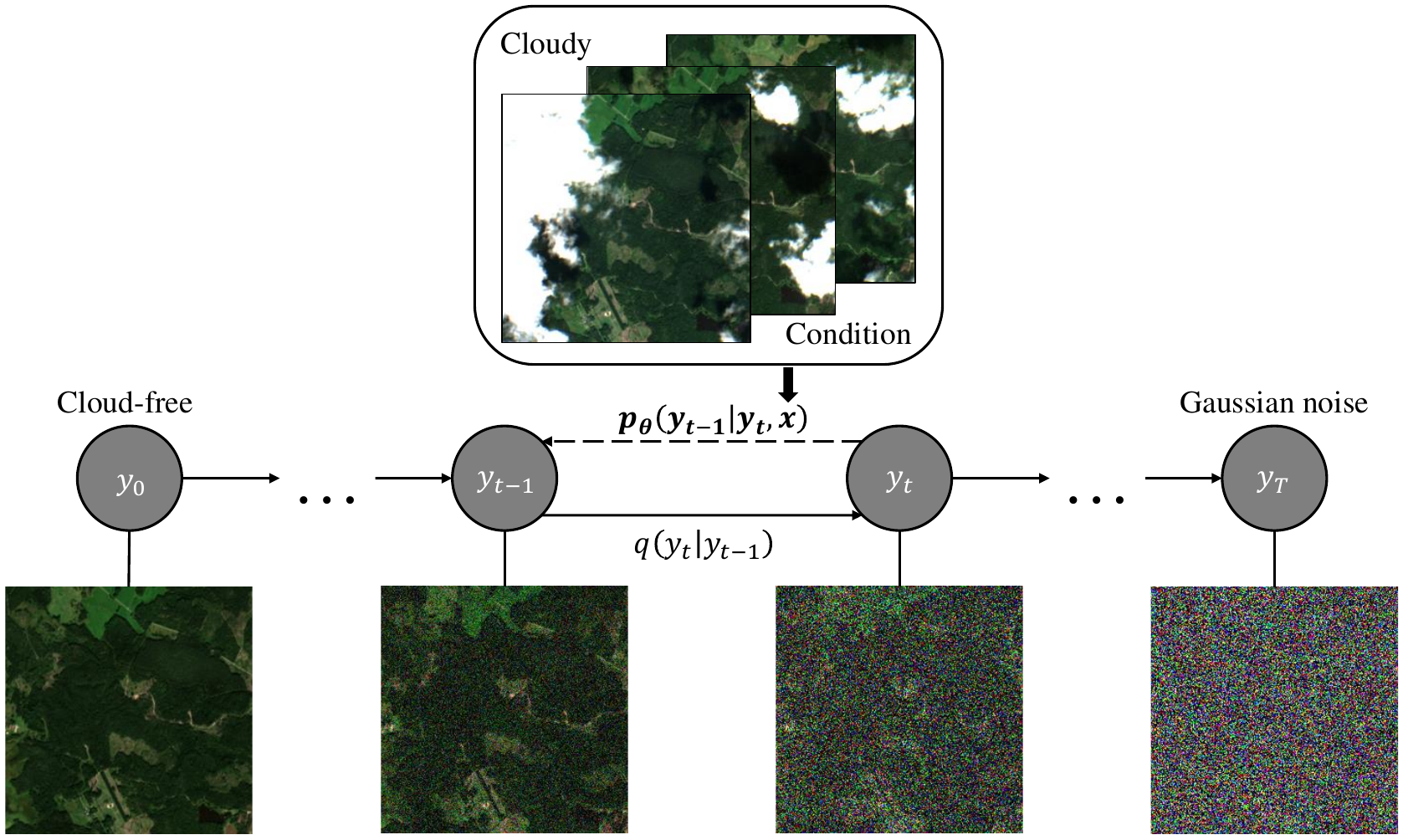}
\caption{Overall pipeline of our proposed cloud removal framework DiffCR. \textbf{1) Training}: this phase consists of the forward process (from left to right) and the reverse process (from right to left). The forward process, also known as the diffusion process, progressively perturbs the ground truth $\mathbf{y}_{0}$ by adding Gaussian noise to control the training of the parameterized neural network $p_{\theta}$ through an intermediate transition distribution. The reverse process, also known as the denoising process, takes the noisy image $\mathbf{y}_{t}$ as input to the $p_{\theta}$ and iteratively refines $\hat{\mathbf{y}_{0}}$. In addition, cloudy images $\mathbf{x}$ and noise level $t$ are conditionally injected into $p_{\theta}$ to guide the generation of cloud-free images. \textbf{2) Inference}: this phase takes the Gaussian noise $\mathbf{y}_{T}$, cloudy images $\mathbf{x}$ and $t$ as input to pre-trained $p_{\theta}$ to iteratively reconstruct the cloud-free image $\hat{\mathbf{y}_{0}}$.}
\label{fig:pipeline}
\end{figure}

\IEEEPARstart{I}{n} recent years, with the rapid advancement of remote sensing technology, satellite imagery, particularly optical satellite images, has found extensive applications across various domains, including agriculture monitoring~\cite{grasslandmonitoring}, ground target detection~\cite{objectdetection1,objectdetection2}, land cover classification~\cite{landcover1,landcover2}. However, cloud interference during the imaging process poses a significant challenge. According to the International Satellite Cloud Climatology Project (ISCCP) data, the global average cloud coverage stands at a staggering 66\%. Analyzing 12 years of observations from the Moderate Resolution Imaging Spectroradiometer (MODIS) on the Terra and Aqua satellites, it was found that an average of 67\% of the Earth's surface is covered by clouds~\cite{background}. The opacity of thick clouds obstructs the underlying object information, drastically reducing their utility and effectiveness, and severely hampers further satellite imagery analysis and application. Therefore, developing an effective method to remove clouds from optical remote sensing images has become an essential and pressing research problem, as a successful cloud removal algorithm would significantly enhance data availability and improve the accuracy of downstream tasks dependent on optical satellite images.

With the advancement of deep learning techniques, generative adversarial networks (GANs) have emerged as a prominent research direction in the field of image processing. GAN-based methods can learn data distributions and generate high-quality samples, finding successful applications in tasks such as image super-resolution~\cite{esrgan,survey-super-resolution,ispa}, image inpainting~\cite{lama}, and style transfer~\cite{stylegan}. In the context of cloud removal, GAN-based methods ~\cite{mcgan,pix2pix,stgan,ctgan,cr-ts-net} have also been widely employed, yielding promising results. Specifically, they consist of two key components: a generator responsible for generating cloud-free surface images from cloud-contaminated images and a discriminator tasked with determining the authenticity of the generated images. The generator and discriminator gradually improve their performance through iterative training, generating more realistic and clear surface images.

Despite the remarkable performance of GAN-based approaches in cloud removal, several challenges remain. For instance, they are inherently difficult to train and can suffer from issues such as mode collapse, leading to distorted or repetitive generated images. Moreover, cloud morphology and distribution exhibit diversity and complexity, placing complex demands on GAN-based methods training and application. Therefore, designing more robust and efficient deep generative models remains crucial in cloud removal research.

To achieve cloud removal while preserving high image fidelity, we propose to explore the applicability of diffusion models for cloud removal in optical satellite images. The diffusion-based approach constructs data labels by a diffusion process with added noise. An iterative refinement process follows this through inverse denoising to produce finer samples. In this manner, rather than modeling complex cloud-free features in one go, our proposed cloud removal framework DiffCR (see Fig.~\ref{fig:pipeline}) decomposes the problem into a series of forward-backward diffusion steps to learn reasonable cloud-free details. In addition, to further improve realism and image quality, we design a novel conditional denoising model composed of three components: 1) condition encoder, 2) time encoder, and 3) denoising autoencoder. The condition encoder is used to learn features of the conditional image, providing a robust color representation to ensure close similarity in appearance information between the conditional input and the synthetic output. The time encoder transforms the time $t$ that reflects the noise level into an implicit time embedding. The denoising autoencoder extracts feature from noisy cloud-free images and fuse them with conditional and time embedding to generate a high-fidelity cloud-free image. Afterward, we further propose an efficient and novel time and condition fusion block (TCFBlock) in the conditional denoising model that explicitly integrates spatial correlation of noisy images, multiscale condition embedding, and implicit time embedding into the diffusion framework. With the aforementioned optimizations and innovations, DiffCR achieves the generation of high-quality samples in a mere single step and complete convergence in 3 to 5 steps. In summary, the contributions of our work can be summarized as follows:

\begin{enumerate}
    \item We present the DiffCR, a novel and fast conditional diffusion framework for cloud removal from optical satellite images. This framework decouples the processing of conditional information, enabling the generation of high-fidelity images in just one sampling step.
    \item We design a novel conditional denoising model with three components: condition encoder, time encoder, and denoising autoencoder. This model ensures a high similarity between the input image and the synthesized output in terms of appearance information and can effectively improve the realism and quality of the image.
    \item We propose an efficient and novel time and condition fusion block (TCFBlock) as the basic unit of the conditional denoising model that can explicitly integrate the multiscale features of noisy images, multiscale condition embedding, and implicit time embedding.
\end{enumerate}

We demonstrate the empirical validity of DiffCR by comparing it with various competitive GAN, regression, and diffusion-based methods on two commonly used benchmark datasets showing SOTA results. Besides, we conduct extensive ablation experiments (see Table~\ref{tab:architecture}) to investigate the impact of various architectural designs in denoising autoencoder. We believe our work will contribute to efficient real-time cloud removal research, achieving high fidelity and robustness, even with limited computational resources, and providing valuable insights for applying conditional diffusion models in other computer vision domains and generative artificial intelligence, such as text-to-image and image-to-image synthesis.

\section{Related Work}\label{sec:relatedwork}

\subsection{Cloud Removal}

\noindent{Cloud removal is an important and challenging research area. Due to weather conditions, cloud cover often obstructs optical satellite imaging, rendering some geographic features invisible and hindering the application of remote sensing images in downstream tasks. Cloud removal aims to restore the obscured geographical information, unlocking the full utilization potential of remote sensing imagery.}

Based on the input data, cloud removal methods can be categorized into single-temporal and multi-temporal approaches. Due to technological limitations, early cloud removal methods~\cite{mcgan,8803666,rice,spa-gan,accv,yuv-color} based on optical satellite imagery primarily focused on single-temporal approaches, aiming to remove clouds and restore land features from a single cloud-contaminated image at a specific geographic location. However, obtaining a cloud-free image from a single frame becomes a significant challenge when the cloud coverage is extensive, as there is insufficient information to facilitate image reconstruction.

With advancements in remote sensing technology, satellites can capture images of the same location at shorter intervals. This advancement lets us easily acquire multiple satellite images captured simultaneously, providing ample data support for designing effective cloud removal methods. Generally, cloud positions vary over time, and the regions obscured by clouds at the same geographic location do not entirely overlap at different moments. Leveraging spectral and temporal information, existing multi-temporal cloud removal methods~\cite{ae,stnet,stgan,ctgan,pmaa} fuse multiple cloudy images to generate detail-rich cloud-free images.

Existing cloud removal approaches have predominantly relied on GANs, enhancing image reconstruction by introducing adversarial losses during training. MCGAN~\cite{mcgan} extended CGANs from RGB images to multi-spectral images for cloud removal. Spa-GAN~\cite{spa-gan} introduced spatial attention mechanisms in GANs to improve information recovery in cloud regions, resulting in higher-quality cloud-free images. AE~\cite{ae} employed convolutional autoencoders trained on multi-temporal remote sensing datasets for cloud removal. STNet~\cite{stnet} integrated cloud detection techniques and fused spatiotemporal features from multiple cloudy images for cloud removal. STGAN~\cite{stgan} regarded cloud removal as a conditional image synthesis problem and proposed a spatiotemporal generative network. CTGAN~\cite{ctgan} introduced a Transformer-based GAN for cloud removal. PMAA~\cite{pmaa} achieved efficient cloud removal using progressive autoencoders. Additionally, there are also some works~\cite{dsen2-cr,cr-ts-net,uncrtaints} that utilize radar imagery to assist in cloud removal, as well as a limited number of works~\cite{ddpm-cr,seqdms} that employ diffusion models. However, due to the limitations of GAN frameworks and adversarial learning mechanisms, existing GAN-based methods struggle to attain satisfactory image fidelity. Moreover, although a few diffusion-based methods have performed well, they are often highly inefficient, typically requiring thousands of sampling steps. Therefore, this work explores a new and fast cloud removal framework using conditional diffusion models, achieving superior fidelity without bells and whistles.

\subsection{Diffusion Models}

\noindent{Image generation has been a popular research direction in computer vision, with various image generation paradigms~\cite{gan,vae,flow} proposed and applied to this task. In recent years, numerous GAN-based models have dominated the state-of-the-art. However, this paradigm has been disrupted by the advent of diffusion models.}

Inspired by thermodynamics~\cite{thermodynamics}, diffusion models gradually introduce noise through iterations, aiming to learn the information decay caused by noise and generate images based on the learned patterns. Diffusion models have achieved outstanding performance in high-resolution image generation~\cite{ddpm}, surpassing GANs in terms of image synthesis quality and diversity~\cite{guided-diffusion}. The success of diffusion models has attracted significant attention from researchers, leading to further improvements in generation quality~\cite{improved,cdm} and sampling speed~\cite{ddim,plms,dpm-solver}.

Recently, conditional diffusion models~\cite{guided-diffusion,classifier-free,ldm,sdm,diffpose,vq-diffusion,glide,dalle,dalle2,imagen,l2sb,sr3,palette,repaint} have been developed and applied to various downstream tasks. Numerous diffusion-based methods for image-to-image translation~\cite{sr3,palette,l2sb,repaint,ldm} have been proposed, targeting low-level visual tasks such as colorization, denoising, inpainting, and super-resolution. Cross-modal works have also demonstrated the remarkable performance of diffusion models in text-to-image generation~\cite{vq-diffusion,glide,dalle,dalle2,imagen}. Furthermore, diffusion models have shown remarkable results in semantic image synthesis~\cite{sdm}, image segmentation~\cite{segdiff}, video generation~\cite{pvdm}, and pose estimation~\cite{diffpose}, among other specific tasks. Inspired by these works, we pioneer the exploration of using diffusion models to tackle the cloud removal task, proposing various improvement strategies to address the complexities of cloud removal and achieve high-fidelity cloud-free images.

\section{Method}\label{sec:diffcr}

\begin{figure*}[t]
\begin{minipage}[t]{0.495\textwidth}
\begin{algorithm}[H]
  \caption{Training a Conditional Denoising Model} \label{alg:training}
  \small
  \hspace*{0.02in} {\bf Input:} Cloudy images $\mathbf{x}$, noisy cloud-free image $\mathbf{y}_t$, noise level $t$ \\
  \hspace*{0.02in} {\bf Output:} $\hat{\mathbf{y}_0}$ estimated by conditional denoising model $\boldsymbol{f}_\theta$ 
  \begin{algorithmic}[1]
    \Repeat
      \State $(\mathbf{x},\mathbf{y}_0) \sim q(\mathbf{x},\mathbf{y})$
      \State $t \sim \mathrm{Uniform}(\{1, \dotsc, T\})$
      \State $\boldsymbol{\epsilon}\sim\mathcal{N}(\mathbf{0},\mathbf{I})$
      \State Take a gradient descent step on
      \Statex $\qquad \nabla_\theta \left\| \mathbf{y}_0-\boldsymbol{f}_\theta(\sqrt{\bar{\alpha}_t}\mathbf{y}_0 + \sqrt{1 - \bar{\alpha}_t}\boldsymbol{\epsilon},t,\mathbf{x}) \right\|^p_p$ 
    \Until{converged}
  \end{algorithmic}
\end{algorithm}
\end{minipage}
\hfill
\begin{minipage}[t]{0.495\textwidth}
\begin{algorithm}[H]
  \caption{Inference with Iterative Refinement} \label{alg:inference}
  \small
  \hspace*{0.02in} {\bf Input:} Cloudy images $\mathbf{x}$, Gaussian noise $\mathbf{y}_T$, noise level $t$ \\
  \hspace*{0.02in} {\bf Output:} $\mathbf{y}_0$ estimated by pre-trained conditional denoising model 
  \begin{algorithmic}[1]
    \vspace{.04in}
    \State $\mathbf{y}_T \sim \mathcal{N}(\mathbf{0}, \mathbf{I})$
    \For{$t=T, \dotsc, 1$}
      \State $\mathbf{z} \sim \mathcal{N}(\mathbf{0}, \mathbf{I})$ if $t > 1$, else $\mathbf{z} = \mathbf{0}$
       \State $\mathbf{y}_{t-1} = \frac{1}{\sqrt{\alpha_t}}\left(\mathbf{y}_t - \frac{1-\alpha_t}{\sqrt{1-\bar\alpha_t}} \boldsymbol{f}_\theta(\mathbf{y}_t, t, \mathbf{x}) \right) + \sqrt{1-\alpha_t} \mathbf{z}$
    \EndFor
    \State \textbf{return} $\mathbf{y}_0$
    \vspace{.04in}
  \end{algorithmic}
\end{algorithm}
\end{minipage}
\vspace{-1em}
\end{figure*}

\subsection{Problem Definition} \label{sec:definition}

\noindent{Cloud removal aims to generate a cloud-free image from cloudy optical satellite images. We define the cloudy images as $\mathbf{x}\in \mathbb{R}^{N \times C \times H \times W}$, where $C$, $H$, and $W$ represent the number of channels, height, and width of the satellite image, respectively, and $N$ represents the number of images. Specifically, when $N=1$, indicating only one cloudy image is input, we refer to it as single-temporal cloud removal; when $N>1$, the input consists of multiple cloudy images acquired at different times but at the same geographical location, and we refer to it as multi-temporal cloud removal. The cloud-free images corresponding to each area are denoted as $\mathbf{y} \in \mathbb{R}^{1\times C \times H \times W}$.}

\subsection{Overall Framework} \label{sec:pipeline}

\noindent{In this paper, we propose a novel framework, called DiffCR, combined conditional diffusion models~\cite{ddpm} and deep convolutional network for cloud removal. As shown in Fig.~\ref{fig:pipeline}, it consists of two parts: the noise injection process from left to right and the denoising process from right to left. Given the multiple cloudy images $\mathbf{x}\in \mathbb{R}^{N \times C \times H \times W}$ as a condition and the degraded image $\mathbf{y} \in \mathbb{R}^{1\times C \times H \times W}$ with a certain noise level $t$ added to the ground truth image $\mathbf{y}_{0}$, our framework can iteratively refine and generate high-quality cloud-free images with fine-grained details.}

\subsubsection{Forward Diffusion} \label{sec:forward}

Forward diffusion is a deterministic process that does not require learning. It follows the Markov assumption and progressively adds noise to the cloud-free image $\mathbf{y}_{0}$ over $T$ iterations:

\begin{equation}
    q(\mathbf{y}_{1:T}|\mathbf{y}_{0})=\prod_{t=1}^{T}q(\mathbf{y}_{t}|\mathbf{y}_{t-1}),
\end{equation}

\begin{equation}
    q(\mathbf{y}_{t}|\mathbf{y}_{t-1})=\mathcal N(\mathbf{y}_{t};\sqrt{1-\beta_{t}}\mathbf{y}_{t-1},\beta_{t}\mathbf{I}),
\end{equation}
where $\beta_{t}$ is a predefined hyperparameter representing the noise variance. Typically, $\beta_{t}$ ranges from 0 to 1, and satisfies $\beta_1<\beta_2<\cdots<\beta_T$. As $T$ increases, $\mathbf{y}_{T}$ gradually loses its original data characteristics. When $T\rightarrow\infty$, $\mathbf{y}_{T}$ becomes a random noise that follows a Gaussian distribution. We refer to the variances at different iterations as the noise schedule.

\begin{figure}[!t]
\centering
\includegraphics[width=\linewidth]{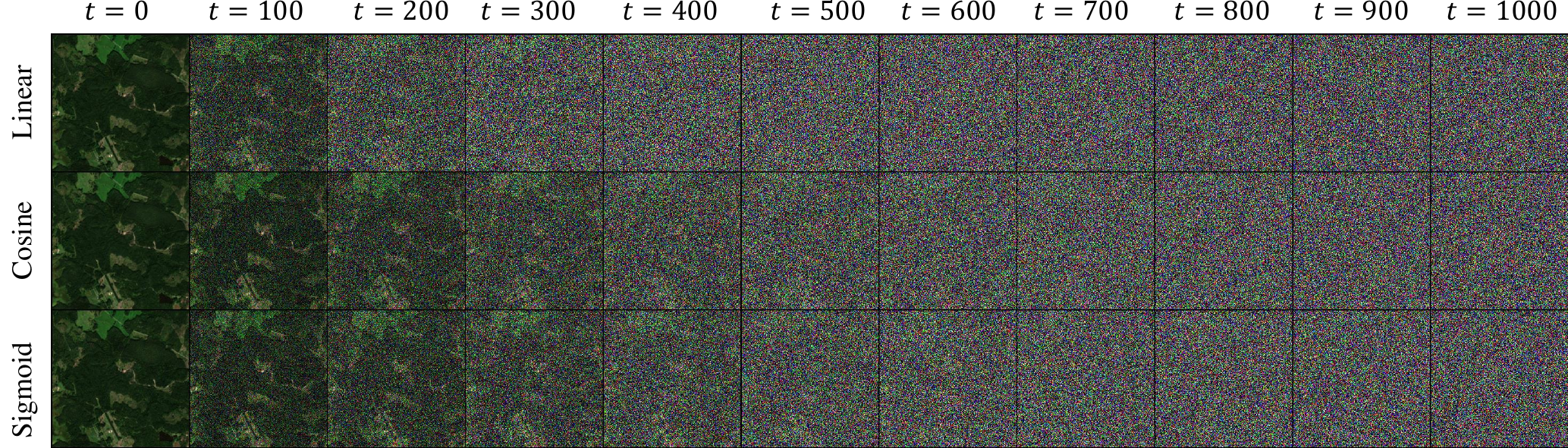}
\caption{Samples generated by the diffusion process with different noise schedules. The linear~\cite{ddpm} schedule produces samples that are almost pure noise in the end, with little difference. In contrast, the cosine~\cite{cosine} and sigmoid~\cite{sigmoid} schedules slowly add noise and generate high-quality samples, which are more beneficial for training the diffusion-based denoising model.}
\label{fig:schedula}
\end{figure}

In fact, through mathematical derivation, we can directly sample $\mathbf{y}_t$ at any step $t$ based on the original data $\mathbf{y}_0$:

\begin{equation}\label{equ:property}
    q(\mathbf{y}_{t}|\mathbf{y}_{0}) = \mathcal N(\mathbf{y}_{t};\sqrt{\overline{\alpha}_{t}}\mathbf{y}_{0}, (1-\overline{\alpha}_{t})\mathbf{I}),
\end{equation}
where $\alpha_{t}=1-\beta_{t}$, and $\overline{\alpha}_{t}=\prod_{i=1}^{t}\alpha_{i}$. This is a crucial property, and from Equation~\eqref{equ:property}, we can see that $\mathbf{y}_{t}$ is actually a linear combination of $\mathbf{y}_{0}$ and Gaussian noise. Furthermore, we can define the noise schedule based on $\overline{\alpha}_{t}$ instead of $\beta_{t}$, such as~\cite{cosine,sigmoid}. An effective noise schedule can make the diffusion process more natural, accelerate model convergence, and improve performance. Fig.~\ref{fig:schedula} illustrates the visual effects of different noise schedule strategies in the diffusion process. 
In Section~\ref{sec:schedule}, we validate the impact of different noise schedules on cloud removal performance.

\subsubsection{Reverse Denoising}\label{sec:reverse}

If we can reverse the diffusion process, \textit{i.e.}, the sample from $q(\mathbf{y}_{t-1}|\mathbf{y}_{t})$, we can reconstruct an accurate cloud-free sample from a random Gaussian distribution $\mathcal N(0, \mathbf{I})$. However, since it is not feasible to directly predict $q(\mathbf{y}_{t-1}|\mathbf{y}_{t})$ from the complete dataset, we generally use a parameterized neural network model $p_{\theta}$ to approximate the conditional probability distribution and perform the reverse denoising process (Algorithm~\ref{alg:training}).

\begin{equation}
    p_{\theta}(\mathbf{y}_{0:T})=p(\mathbf{y}_{T})\prod_{t=1}^{T}p_{\theta}(\mathbf{y}_{t-1}|\mathbf{y}_{t},\mathbf{x}),
\end{equation}

\begin{equation}
    p_{\theta}(\mathbf{y}_{t-1}|\mathbf{y}_{t},\mathbf{x})=\mathcal N(\mathbf{y}_{t-1};\boldsymbol{\mu}_{\theta(\mathbf{y}_{t},t,\mathbf{x}),\boldsymbol{\Sigma}_{\theta}(\mathbf{y}_{t},t,\mathbf{x})}),
\end{equation}
where $\mathbf{y}_{t}$, $t$, and $\mathbf{x}$ are inputs to the model, and $\boldsymbol{\mu}_{\theta}$ and $\boldsymbol{\Sigma}_{\theta}$ represent the mean and variance of the prior distribution predicted by the model.

It is noteworthy that inference (Algorithm~\ref{alg:inference}) is also a reverse Markov process, but it starts from Gaussian noise and samples forward step by step, \textit{i.e.}, it generates high-fidelity cloud-free images through iterative refinement. Theoretically, the larger the time step $T$ for adding noise, the higher the quality of the final inferred image generated. Existing diffusion models~\cite{palette,sr3,sdm} typically adopt the sampling schemes proposed by~\cite{ddpm} or~\cite{ddim}, often requiring thousands of steps to generate high-quality images, which severely affects the practical application of the model, especially on edge devices with low resources. To address this challenge, we introduce a fast ODE solver~\cite{dpm-solver}\footnote{Please refer to the supplementary material for code implementation details.} that can generate high-quality images in about 10 steps. 
In the ablation study Section~\ref{sec:sampling}, we provide quantitative and qualitative comparisons of the results from different sampling steps of various samplers.

\begin{figure*}[!t]
\centering
\includegraphics[width=\linewidth]{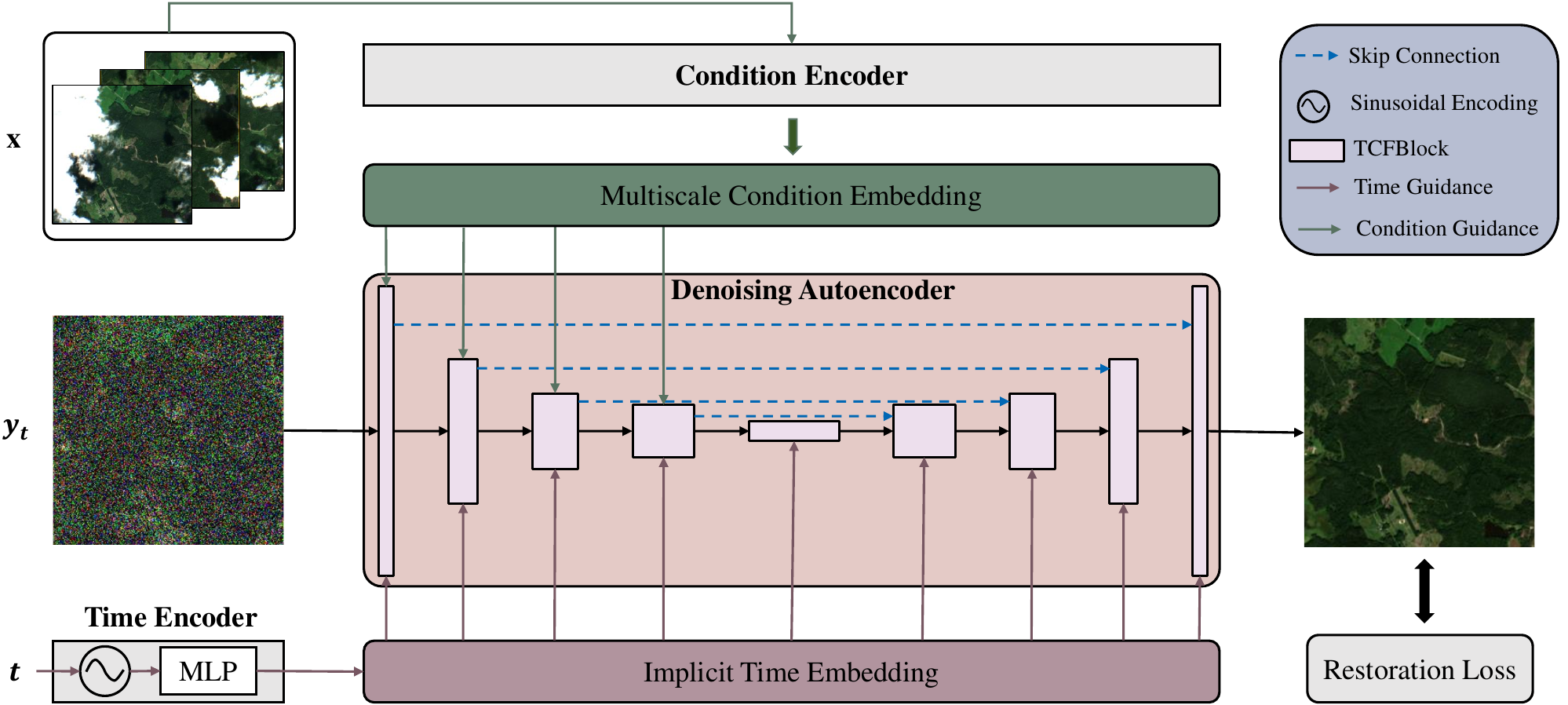}
\caption{Overall architecture of the conditional denoising model of DiffCR. The model consists of three parts: 1) a condition encoder, 2) a time encoder, and 3) a denoising autoencoder. The condition encoder and time encoder are responsible for extracting the spatial features of cloudy images $\mathbf{x}$ and temporal features of noise level $t$, respectively, which are then fed into the denoising autoencoder to guide the entire denoising process. The condition encoder and denoising autoencoder comprise several of our developed TCFBlocks (refer to Fig.~\ref{fig:tcfblock}). The time encoder comprises a sinusoidal encoding function and a multilayer perceptron (MLP). Ultimately, the data distribution of cloud-free images $\mathbf{y_0}$ is estimated under the supervision of the restoration loss.}
\label{fig:unet}
\end{figure*}

\subsection{Conditional Denoising Model}\label{sec:cdm}

\noindent{We present a novel conditional denoising model based on our carefully designed time and condition fusion block (TCFBlock) as the fundamental component, as illustrated in Fig.~\ref{fig:unet}. The model consists of three parts: 1) a condition encoder, 2) a time encoder, and 3) a denoising autoencoder. The condition encoder and time encoder are responsible for extracting the multiscale spatial features of cloudy images $\mathbf{x}$ and temporal features of noise level $t$, respectively, which are then fed into the denoising autoencoder to guide the entire reconstruction process. Under the supervision of the restoration loss, the model outputs high-fidelity cloud-free images.}

\subsubsection{Time and Condition Fusion Block}\label{sec:tcfblock}

\begin{figure}[!t]
\centering
\includegraphics[width=\linewidth]{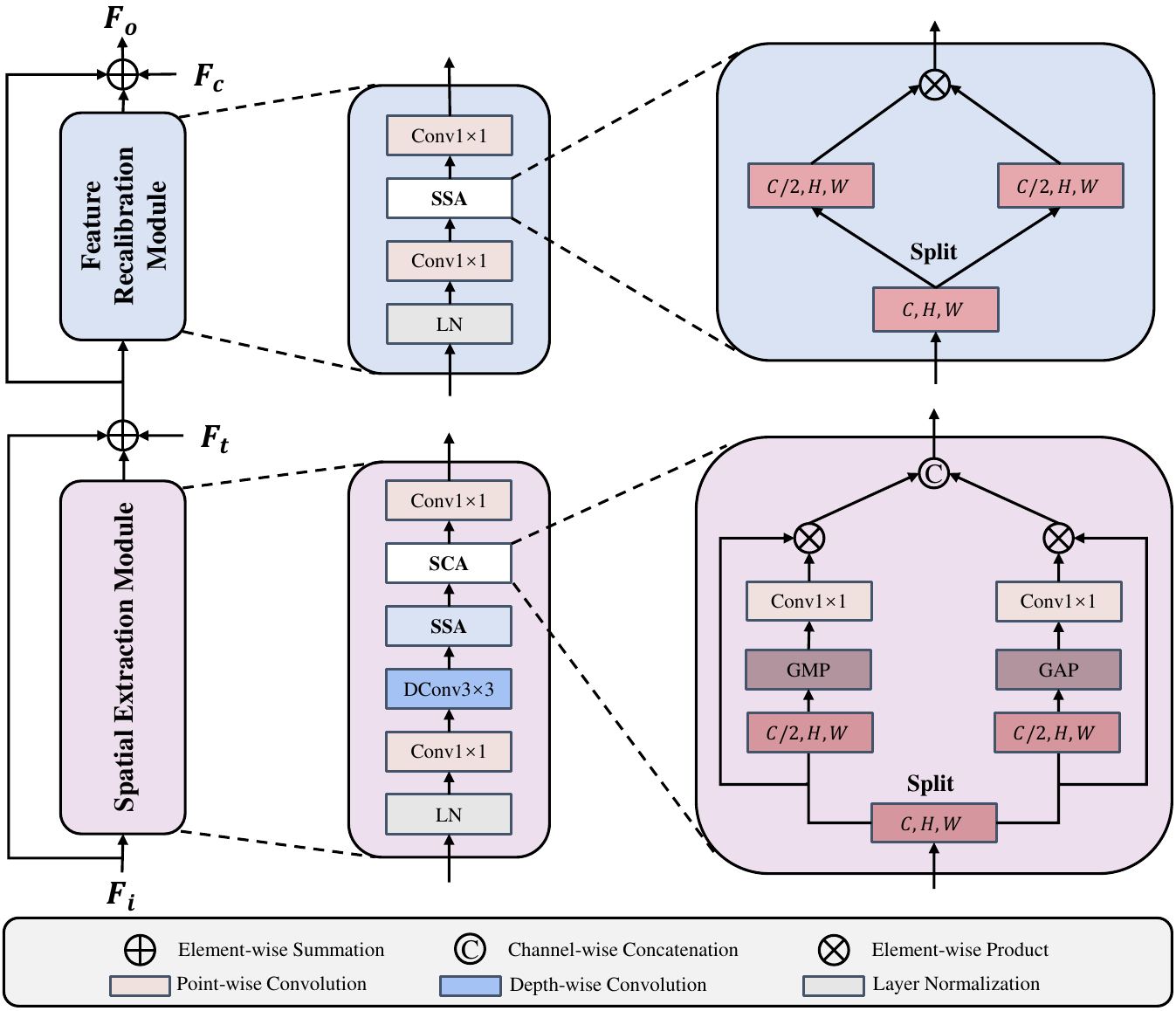}
\caption{Schematic diagram of our proposed time and condition fusion block.}
\label{fig:tcfblock}
\end{figure}

It is well-known that the training of diffusion models typically involves many iterative steps. These require substantial computational resources. Motivated by this and inspired by the network structures in~\cite{resnet,guided-diffusion,nafnet,convnext}, we design an efficient time and condition fusion block (TCFBlock) as the basic unit of the conditional denoising model. It significantly reduces computational complexity while achieving SOTA performance. As illustrated in Fig.~\ref{fig:tcfblock}, TCFBlock comprises the spatial extraction module (SEM) and the feature recalibration module (FRM). 

\begin{equation}
    \mathbf{F}_n = \operatorname{SEM}(\mathbf{F}_i)+\mathbf{F}_i+\mathbf{F}_t,
\end{equation}

\begin{equation}
    \mathbf{F}_o = \operatorname{FRM}(\mathbf{F}_n)+\mathbf{F}_n+\mathbf{F}_c,
\end{equation}
where $\mathbf{F}_i \in \mathbb{R}^{C \times H \times W}$ represents the input features, $\mathbf{F}_n \in \mathbb{R}^{C \times H \times W}$ represents the output of the feature by SEM, $\mathbf{F}_c \in \mathbb{R}^{C \times H \times W}$ represents the spatial features of cloudy images, $\mathbf{F}_t \in \mathbb{R}^{C \times 1 \times 1}$ represents the temporal features of noise level, and $\mathbf{F}_o \in \mathbb{R}^{C \times H \times W}$ is the final output features through FRM.

\paragraph{Spatial Extraction Module} SEM is designed to extract fine-grained spatial features and is constructed as an inverted bottleneck structure. Firstly, layer normalization~\cite{ln} commonly used in Transformers~\cite{transformer,vit,swin} is employed to replace batch normalization~\cite{bn} for normalizing input features, aiming to stabilize training. Then, a $1\times1$ convolution with channel expansion and a depthwise separable convolution with a $3\times3$ kernel are utilized to extract detailed spatial features. Subsequently, a split self-activation (SSA) mechanism~\cite{nafnet} is introduced to replace the role of conventional activation functions, thereby enhancing the non-linear expressive power of the model.

\begin{equation}\label{equ:spational}
    \mathbf{F}_{g}=\operatorname{SSA}\left(\operatorname{DConv}_{3\times3}\left(\operatorname{Conv}_{1\times1}\left(\phi\left(\mathbf{F}_{i}\right)\right)\right)\right),
\end{equation} 
where $\phi$ represents normalization. $\operatorname{Conv}$ and $\operatorname{DConv}$ respectively denote standard and depthwise separable convolution.

In recent years, channel attention mechanisms~\cite{senet,sknet,cbam} have achieved great success in various computer vision tasks. Therefore, we introduce a novel and efficient split channel attention (SCA) into SEM to enhance the robustness of the extracted features. As shown in Fig.~\ref{fig:tcfblock}, SCA comprises two channel attention branches: the global channel attention branch and the local channel attention branch. Firstly, the input features are divided into two parts in the channel dimension, followed by global average pooling (GAP) and global maximum pooling (GMP) in each part, respectively. Subsequently, a $1\times1$ convolution is employed to establish explicit correlations between the channels in each part. Finally, the selected features from the two branches are concatenated along the channel dimension to generate robust features that simultaneously consider global and local information. As average pooling emphasizes the extraction of mean information of features and can retain some intricate details of the image, we refer to the branch where it is located as the global channel attention branch. Conversely, since maximum pooling focuses more on extracting the most salient features of the image, we denote the branch where it is located as the local channel attention branch. Lastly, a $1\times1$ convolution is introduced to modulate the concatenated features further.
\begin{equation}\label{equ:sca}
    \mathbf{F}_{s}=\mathbf{W}_0(\operatorname{SCA}(\mathbf{F}_g)),
\end{equation} 
\begin{equation}\label{equ:sca}
    \operatorname{SCA}(\mathbf{F}_g)=\Phi(\mathbf{W}_1(\operatorname{GAP}(\mathbf{F}_{g1})), \mathbf{W}_2(\operatorname{GMP}(\mathbf{F}_{g2}))),
\end{equation} 
where $\mathbf{F}_{g1} \in \mathbb{R}^{\frac{C}{2} \times H \times W}$ and $\mathbf{F}_{g2} \in \mathbb{R}^{\frac{C}{2} \times H \times W}$ are the results of splitting $\mathbf{F}_g \in \mathbb{R}^{C \times H \times W}$ in the channel dimension. $\mathbf{W}_0 \in \mathbb{R}^{C \times C}$, $\mathbf{W}_1 \in \mathbb{R}^{\frac{C}{2} \times \frac{C}{2}}$, and $\mathbf{W}_2 \in \mathbb{R}^{\frac{C}{2} \times \frac{C}{2}}$ respectively denote the weight matrices of the pointwise convolution. $\Phi$ represents channel-wise concatenation operation.


\paragraph{Feature Recalibration Module} FRM is designed to recalibrate the spatial features of the conditional and noisy images, the temporal features of the noise level, which consists of two pointwise convolution layers, one activation layer, and one normalization layer (Equation~\eqref{equ:cfm}). To reduce computational complexity, we adopt SSA~\cite{nafnet} to replace the commonly used ReLU~\cite{relu} or SiLU~\cite{silu} activation functions in deep models. In~\cite{convnext}, it was found that fewer activation functions can lead to better performance. Therefore, we employ only a single SSA to enhance the model's nonlinearity. Additionally, the inverted bottleneck structure~\cite{mobilenetv2}, which has larger middle and smaller ends, has been proven to prevent information loss effectively. Hence, we use it in the two pointwise convolutions instead of the bottleneck structure~\cite{resnet}.

\begin{equation}\label{equ:cfm}
    \mathbf{F}_{r}=\mathbf{V}_2(\sigma(\mathbf{V}_1(\phi(\mathbf{F}_{m}))),
\end{equation}
where $\mathbf{F}_r \in \mathbb{R}^{C \times H \times W}$ represents the FRM recalibrated features, $\sigma$ denotes the activation function, and $\phi$ refers to layer normalization. $\mathbf{V}_1 \in \mathbb{R}^{C \times 2C}$ and $\mathbf{V}_2 \in \mathbb{R}^{2C \times C}$ respectively denote the weight matrices of the two pointwise convolutions in the inverted bottleneck structures.

\subsubsection{Condition Encoder}

Existing conditional diffusion models~\cite{palette,sr3} directly concatenate the conditional input and noisy input along the channel dimension and feed them into the denoising model, which increases the difficulty of model training. Motivated by this, we develop a lightweight decoupled CNN-based condition encoder, which is composed of several TCFBlock (without $\mathbf{F}_t$ and  $\mathbf{F}_c$ injection) followed by downsampling layers to explicitly extract multi-scale features of the conditional image $\mathbf{x}$. Specifically, for the conditional input $\mathbf{x}$, we first utilize the TCFBlock to refine the spatial relationship and then perform scale reduction. Following~\cite{swin}, we introduce convolutional layers with kernel and stride size of $2\times2$ for downsampling, and to mitigate inherent information loss during downsampling, we increase the number of channels.

\begin{equation}\label{cencoder}
    \mathbf{F}_c= \operatorname{DownSample}(\operatorname{TCFBlock}(\mathbf{x})),
\end{equation}
where $\operatorname{DownSample}(\cdot)$ refers to the convolution mentioned above, and $\mathbf{x}$ and $\mathbf{F}_c$ denote the input conditional image and the output feature after scale reduction, respectively.

\subsubsection{Time Encoder}

To prevent the denoising autoencoder from learning a fixed paradigm, we choose to add images with varying levels of random noise during input. Correspondingly, we need to include the time step $t$ that reflects the level of noise as an additional input for the denoising autoencoder to guide the entire learning process. Specifically, for the time step $t$, we adopt the same time encoder~\eqref{equ:tencoder} as in~\cite{ddpm,guided-diffusion,palette,sr3}. First, we perform a sinusoidal encoding and feed it to a multi-layer perceptron to obtain an implicit time embedding. Finally, after broadcasting its dimensions, the time encoding is injected into each TCFBlock of the denoising autoencoder. Since the time step $t$ does not contain spatial information, convolution operations similar to those used for images are unnecessary. These operations fail to bring additional performance gains and increase computational overhead.

\begin{equation}\label{equ:tencoder}
    \mathbf{F}_t = \operatorname{Reshape}(\operatorname{MLP}(\sin(t))),
\end{equation}
where $t$ represents the time step, $\sin(\cdot)$ denotes the sinusoidal encoding, $\operatorname{MLP}$ represents a multilayer perceptron consisting of two linear layers with reverse bottleneck structure and a SiLU~\cite{silu} activation function. $\operatorname{Reshape}(\cdot)$ denotes broadcasting the time embedding to modulate weights of image features, and $\mathbf{F}_t$ represents implicit time embedding.

\subsubsection{Denoising Autoencoder}

We utilize the TCFBlock in Section~\ref{sec:tcfblock} as the basic unit to design a UNet\cite{unet}-structured autoencoder for the denoising process of DiffCR.

\paragraph{Encoder} The encoder comprises an array of consecutively stacked TCFBlocks and several downsampling layers. This design aims to extract diverse multi-scale features at varying resolutions, encapsulated by Equation~\eqref{equ:encoder}. Informed by~\cite{swin}, we adopt a downsampling scheme using convolutional layers characterized by a $2 \times 2$ kernel size and stride. To mitigate the inherent information loss during the downsampling operation, we simultaneously expand the number of channels, which essentially enhances the feature representation capacity of each layer.

\begin{equation}\label{equ:encoder}
    \mathbf{F}_{e+1} = \operatorname{DownSample}(\operatorname{TCFBlock}(\mathbf{F}_e, \mathbf{F}_c, \mathbf{F}_t)),
\end{equation}
where $\mathbf{F}_e \in \mathbb{R}^{C \times H/e \times W/e}$ and $\mathbf{F}_{e+1} \in \mathbb{R}^{C \times H/(e+1) \times W/(e+1)}$ respectively denote the feature maps extracted from the $e$-th and $(e+1)$-th layers. $DownSample(\cdot)$ operation embodies the convolutional downsampling approach, further enhancing the hierarchical representation of our model. By progressively accumulating abstract features layer by layer, our encoder provides a robust and rich feature set that boosts the performance of downstream tasks.

\paragraph{Middle}

The purpose of the middle block is to modulate the features extracted by the encoder at the minimum spatial scale, further refining the features under the guidance of a large receptive field or establishing spatial positional correlations among the features. Therefore, we can inherit the encoder's design by stacking multiple TCFBlocks continuously.

\begin{equation}\label{equ:middle}
    \mathbf{F}_{m+1} = \operatorname{TCFBlock}(\mathbf{F}_m, \mathbf{F}_c, \mathbf{F}_t).
\end{equation}

\paragraph{Decoder}


The design of the decoder is similar to the encoder, where multiple TCFBlocks are stacked, and upsampling layers are utilized to gradually restore the latent features to the spatial scale of the original input image. To mitigate information loss, we leverage PixelShuffle~\cite{pixelshuffle} as a replacement for interpolation or transposed convolution for upsampling. Similar to UNet~\cite{unet}, we also introduce skip connections. However, instead of using the standard channel-wise concatenation operation, we directly use element-wise addition. The skip connections with direct addition do not introduce additional parameters, whereas channel-wise concatenation requires learning additional parameters to adjust the dimensionality of the feature channels. When the low-level features are directly added to the high-level features, the low-level features can be more directly fused with the high-level features, while channel-wise concatenation may require more complex transformations to achieve feature fusion. 

\begin{equation}\label{equ:decoder}
    \mathbf{F}_{d+1} = \operatorname{TCFBlock}(\operatorname{UpSample}(\mathbf{F}_d)+\mathbf{F}_e, \mathbf{F}_c, \mathbf{F}_t).
\end{equation}

\subsection{Regression Target and Loss Function}\label{sec:regression-loss}

\noindent{The existing unconditional diffusion models~\cite{ddpm} and conditional diffusion models~\cite{guided-diffusion,classifier-free,palette,sr3,sdm} commonly adopt the noise prediction approach by regressing the output of $f_\theta$ to $\mathbf{\epsilon}$, wherein the neural network model is trained to learn the Gaussian noise added at each step of the forward process.}

\begin{equation}
    \mathcal{L}_{t-1}^{\text{simple}}=\mathbb{E}_{\mathbf{y}_{0},\mathbf{\epsilon}}\| \mathbf{\epsilon}- f_\theta(\sqrt{\bar{\alpha}_t}\mathbf{y}_0 + \sqrt{1 - \bar{\alpha}_t}\mathbf{\epsilon},t,\mathbf{x})\|_p^p,
\end{equation}
where $f_\theta$ represents the parameterized denoising model, $\mathbf{\epsilon}\sim \mathcal{N}(\mathbf{0}, \mathbf{I})$ represents randomly sampled noise from a Gaussian distribution, and $\mathbf{x}$ and $\mathbf{y}$ represent samples from the dataset.  

However, we observe that the noise prediction method increases training difficulty and can even lead to mode collapse when there is abundant conditional input information. Motivated by this consideration, we propose a novel and more straightforward prediction approach by regressing the output of $f_\theta$ to $\mathbf{y}_0$, termed ``data prediction'', as a replacement for noise prediction. 

\begin{equation}
    \mathcal{L}_{t-1}^{\text{simple}}=\mathbb{E}_{\mathbf{y}_{0},\mathbf{\epsilon}}\| \mathbf{y}_0- f_\theta(\sqrt{\bar{\alpha}_t}\mathbf{y}_0 + \sqrt{1 - \bar{\alpha}_t}\mathbf{\epsilon},t,\mathbf{x})\|_p^p.
\end{equation}

Although the values of $\mathbf{\epsilon}$ and $\mathbf{y}_0$ can be determined by each other, changing the regression target affects the scale of the loss function, thereby influencing model training. Since the conditional input image and the cloud-free ground truth image share more similar data distributions, data prediction is significantly easier than noise prediction, resulting in improved model convergence and cloud removal performance. This point is further validated through ablation experiments in Section~\ref{sec:predicion}. We compare these two different regression methods and find that directly predicting $\mathbf{y}_0$ (our proposed data prediction) leads to easier convergence, significantly when the number of condition $\mathbf{x}$ is increased. It is worth noting that DALL·E 2~\cite{dalle2} developed by OpenAI also reports a similar finding in their work.


\section{Experiments}

\subsection{Datasets}

\noindent{We perform all experiments on the following large-scale multi-temporal cloud removal datasets for optical satellite images: the Sen2\_MTC\_Old~\cite{stgan} and the Sen2\_MTC\_New~\cite{ctgan}.}

\subsubsection{Sen2\_MTC\_Old}

The dataset is the first to address the challenge of cloud removal and consists of multi-temporal optical satellite (Sentinel-2) images. It contains 945 tiles from various locations worldwide, with each tile costing $10980\times10980$ pixels and a ground resolution of 10m/pixel. On average, new images are captured every six days at the same location. Subsequently, each tile is cropped into multiple patches of size $256\times256$ pixels. The patches with cloud coverage below 1\% are labeled as ``clear,'' those with cloud coverage ranging from 10\% to 30\% are labeled as ``cloudy,'' and patches with cloud coverage exceeding 30\% are discarded. For the same location, three adjacent cloudy images are selected as inputs, while the clear image with the least cloud coverage is used as the label. In total, 3130 image pairs are obtained for the multi-temporal cloud removal task.

\subsubsection{Sen2\_MTC\_New}

To address the issues of poor annotation quality and low sample resolution in Sen2\_MTC\_Old, \textit{Huang et al.} constructed a new Sentinel-2 satellite image dataset for training multi-temporal cloud removal models. This dataset contains approximately 50 non-overlapping tiles, and each tile consists of around 70 pairs of cropped $256\times256$ pixel patches. Similar to Sen2\_MTC\_Old, Sen2\_MTC\_New follows the same procedure to construct multi-temporal image pairs, where three cloudy images at the same location correspond to one cloud-free image. Specifically, Sen2\_MTC\_Old compressed the images, with pixel values ranging from 0 to 255, while Sen2\_MTC\_New preserved the images in their original TIFF format, with the pixel value ranging from 0 to 10000.

\subsection{Implementation Details}

\noindent{To facilitate experimental reproducibility, this section presents a detailed account of the hyperparameter configurations employed during the training process and the metrics utilized to evaluate the experimental outcomes.}

\subsubsection{Training Settings}

We conducted extensive experiments to validate the efficacy of DiffCR. For the optimization algorithm, we selected AdamW~\cite{adamw} as our optimizer with an initial learning rate of 5e-5 and weight decay of 0. During the training phase, we employed batch sizes of 8. The training process was conducted over a total of about 7M steps. The weight initialization in this work employs Kaiming normal distribution~\cite{kaiming} to enable the network to converge faster and achieve better performance during training. An exponential moving average (EMA)~\cite{ema} is applied to the model parameters, starting from the first iteration of training and updated after each training iteration with a decay rate of 0.9999, which helps stabilize the training process and prevent overfitting. During diffusion progress, we set the maximum noise step to 2000. Experiments were conducted on a workstation equipped with two AMD EPYC 7402 24-core processors, 128 GB of memory, and four NVIDIA GeForce RTX 3090 GPUs with 24 GB of memory. The operating system used was Ubuntu 22.04, and all networks were implemented in PyTorch 1.10.1 with CUDA 11.1 and Python 3.8.

\subsubsection{Evaluation Metrics}

In all experiments, we report the Peak Signal-to-Noise Ratio (PSNR, dB), Structural Similarity Index Measure (SSIM~\cite{ssim}), Learned Perceptual Image Patch Similarity (LPIPS~\cite{lpips}) and Frechet Inception Distance (FID~\cite{fid}) of the test set to evaluate the quality of the generated cloud-free images. It is noteworthy that PSNR and SSIM measure the differences between images on a pixel basis, while FID and LPIPS measure the differences between images based on deep feature vectors. Compared to traditional image quality assessment metrics (PSNR and SSIM), FID and LPIPS align more with human visual perception. Moreover, LPIPS can be almost considered a regional FID metric, as even when FID varies greatly, LPIPS scores show minimal changes. Therefore, when LPIPS approaches proximity, we prioritize FID. For the evaluation of model efficiency, we report the number of parameters (Params, M) and multiply–accumulate operations (MACs, G) for all models, which are calculated via the open source tool\footnote{\url{https://github.com/Lyken17/pytorch-OpCounter}}.

\begin{table}[t]
\begin{center}
\caption{Quantitative comparison of the cloud removal performance of noise schedules selected during the diffusion process of DiffCR.}
\label{tab:schedule}
\setlength{\tabcolsep}{9.5pt}
\scalebox{1.00}{
\begin{tabular}{lcccc}
\toprule
\textbf{Schedule} & \textbf{PSNR~$\uparrow$}   & \textbf{SSIM~$\uparrow$}  & \textbf{FID~$\downarrow$}  & \textbf{LPIPS~$\downarrow$}   \\
\midrule
Linear~\cite{ddpm}      & 19.060 & 0.663 & 92.167 & 0.294 \\
Cosine~\cite{cosine}    & 19.148 & 0.665 & 84.940 & 0.293 \\
Sigmoid~\cite{sigmoid}  & \textbf{19.150} & \textbf{0.671} & \textbf{83.162} & \textbf{0.291} \\
\bottomrule
\end{tabular}}
\end{center}
\end{table}
\begin{table}[t]
\begin{center}
\caption{Quantitative comparison of the cloud removal performance between noise prediction and data prediction during the reverse process of DiffCR. $N$ represents the number of conditional cloudy images.}
\label{tab:regression}
\setlength{\tabcolsep}{4pt}
\scalebox{1.00}{
\begin{tabular}{c|c|cccc}
\toprule
\textbf{Regression}                        & $N$ & \textbf{PSNR~$\uparrow$}            & \textbf{SSIM~$\uparrow$}           & \textbf{FID~$\downarrow$}           & \textbf{LPIPS~$\downarrow$}            \\
\midrule
\multirow{3}{*}{Noise Prediction~\cite{ddpm}} & 1 & 3.738                 & 0.215                & 410.835                & 0.924                \\ \cline{2-6} 
                                & 2 & 3.711                & 0.106                & 504.806                & 1.145                \\ \cline{2-6} 
                                & 3 & 3.641                & 0.099                & 511.454                & 1.205               \\ \midrule
\multirow{3}{*}{Data Prediction [Ours]}  & 1 & 15.139                 & 0.433               & 154.243                & 0.547                \\ \cline{2-6} 
                                & 2 & 16.222                  & 0.444                & 150.771                 & 0.484                \\ \cline{2-6} 
                                & 3 & \textbf{19.150} & \textbf{0.671} & \textbf{83.162} & \textbf{0.291} \\ 
\bottomrule
\end{tabular}}
\end{center}
\end{table}
\begin{table}[t]
\begin{center}
\caption{Quantitative comparison of the cloud removal performance of different processing methods of conditional cloudy images during the reverse process of DiffCR.}
\label{tab:condition}
\setlength{\tabcolsep}{8pt}
\scalebox{1.00}{
\begin{tabular}{ccccc}
\toprule
\textbf{Condition Process} & \textbf{PSNR~$\uparrow$}   & \textbf{SSIM~$\uparrow$}  & \textbf{FID~$\downarrow$}   & \textbf{LPIPS~$\downarrow$}   \\
\midrule
Concat~\cite{palette,ddpm-cr} & 18.821 	& 0.638 & 98.387 & 0.342 \\
Decouple~[Ours]      & \textbf{19.150}
	& \textbf{0.671}
	& \textbf{83.162} 
        & \textbf{0.291} \\
\midrule
$\Delta$ & \textcolor[RGB]{3,191,61}{+~0.329} & \textcolor[RGB]{3,191,61}{+~0.033} & \textcolor[RGB]{3,191,61}{-~15.225} & \textcolor[RGB]{3,191,61}{-~0.051} \\
\bottomrule
\end{tabular}}
\end{center}
\end{table}
\begin{table}[t]
\begin{center}
\caption{Quantitative comparison of the cloud removal performance and efficiency of different sampling steps during the inference phase of DiffCR.}
\label{tab:sampling}
\setlength{\tabcolsep}{7pt}
\scalebox{1.00}{
\begin{tabular}{cccccc}
\toprule
\textbf{Step} & \textbf{Latency (s)} & \textbf{PSNR ↑} & \textbf{SSIM ↑} & \textbf{FID ↓}  & \textbf{LPIPS ↓} \\
\midrule
1	 & 0.09 	   & 19.150 & 0.671  & 83.162 & 0.291   \\
\rowcolor[RGB]{217,217,217}
3	 & 0.16 	   & \textbf{19.182} & \textbf{0.674}  & \textbf{82.978} & \textbf{0.291}   \\
5    & 0.23        & 19.164 & 0.674  & 83.281 & 0.291   \\
10   & 0.39        & 19.170 & 0.672  & 83.177 & 0.291   \\
50   & 1.74        & 19.102 & 0.666  & 84.635 & 0.292   \\
100  & 3.39        & 19.077 & 0.664  & 84.998 & 0.292   \\
\bottomrule
\end{tabular}}
\end{center}\vspace{-1.5em}
\end{table}

\subsection{Ablation Studies}\label{sec:ablation}

\noindent{In this section, we present extensive ablation studies to validate the effectiveness of the proposed DiffCR and its constituent parts. Specifically, we investigate the impact of the following components: 1) the noise schedule in the forward diffusion, 2) the regression target of the denoising model, 
3) the processing method for conditional cloudy images, 4) sampling steps during the inference phase, and 5) changes in the architecture of the denoising model. All ablation experiments are conducted on the Sen2\_MTC\_New dataset due to its high annotation quality and superior image resolution compared to the Sen2\_MTC\_Old dataset.}

\subsubsection{Does Noise Schedule Matter}\label{sec:schedule}

The results in Table~\ref{tab:schedule} present the impact of different noise schedules during the diffusion process. It can be observed that the sigmoid schedule achieves the best cloud removal performance, with a slight decrease in performance for the cosine schedule, and the linear schedule exhibits the poorest performance, which is consistent with theoretical expectations described in Section~\ref{sec:forward}. Therefore, we recommend using the sigmoid schedule for diffusion-based cloud removal or other generative tasks, as it helps with model convergence and performance improvement.

\subsubsection{Data Prediction or Noise Prediction}\label{sec:predicion}

We experimentally verified the impact of the proposed data prediction and native noise prediction on cloud removal performance, as shown in Table~\ref{tab:regression}. As expected (described in Section~\ref{sec:regression-loss}), data prediction outperforms noise prediction for any value of $N$ in $\{1, 2, 3\}$, where $N$ represents the sequence length of cloudy images. In particular, as $N$ increases, data prediction performs increasingly better, while noise prediction performs increasingly worse. This is because, in the data prediction, the model attempts to predict a target (\textit{i.e.}, the cloud-free image after denoising) that is more similar in distribution to the input (the noisy cloud-free image and the cloudy image), especially as the similarity of the distribution increases with increasing $N$. In contrast, noise typically has an entirely different distribution (\textit{e.g.}, Gaussian noise). Therefore, for a model that attempts to extract useful information from the input, predicting a target that is similar to the input may be more accessible, while noise prediction often leads to pattern collapse.



\subsubsection{Processing of Conditional Cloudy Images}\label{sec:condition}

To address the conditional cloudy images processing in the DiffCR, we quantitatively verify two different feature extraction and fusion strategies: 1) Concat: the noisy cloud-free image is concatenated with all the cloudy images along the channel dimension and then fed into the denoising model for processing, which is the most straightforward approach adopted by previous works such as \cite{palette,sr3,ddpm-cr,seqdms}; 2) Decouple: it is a decoupling condition encoder used in DiffCR.
The experimental results in Tabel~\ref{tab:condition} demonstrate that ``Decouple" is the better conditional image processing method, achieving a higher cloud removal performance with improvements of 0.329 in PSNR and 15.225 in FID.

\subsubsection{Inference using Different Sampling Steps}\label{sec:sampling}

The vanilla diffusion model~\cite{ddpm} has an unacceptable slow inference speed, requiring thousands of iterations for refining a high-quality image. Even on suitable GPU devices, its inference speed for a single sample can be slower than training for a single epoch, seriously hindering the application of diffusion models in computer vision. The existing diffusion-based cloud removal models~\cite{ddpm-cr,seqdms} do not consider this real-world constriction, requiring many sampling steps during the inference process. In contrast, owing to the design of its framework that leverages data prediction and conditional decoupling, DiffCR can generate high-quality samples in just one step and achieve absolute convergence in 3 to 5 steps. We compare quantitative results regarding sample quality and inference latency at different sampling steps. The results are summarized in Table~\ref{tab:sampling}. It can be observed that the inference latency increases with an increase in the number of sampling steps. Remarkably, excellent results can be obtained with just one sampling step, with marginal performance improvements from 1 to 3 sampling steps and convergence being achieved between 3 to 5 steps. However, as the number of sampling steps exceeds 25 steps, the performance of cloud removal, as measured by metrics like LPIPS, gradually declines due to over-sampling. Therefore, we ultimately compare the results obtained with one sampling step to other methods, which achieve SOTA performance with only 0.09 seconds of inference latency.

\begin{table*}[t]
    \centering
    \caption{Quantitative comparison of the cloud removal performance and efficiency of architecture changes in denoising autoencoder.}
    \label{tab:architecture}
    \setlength{\tabcolsep}{3.5pt}
    \scalebox{1.00}{
    \begin{tabular}{c|c|ccc|c|cccc|cc}
    \toprule
                            &                                                                        & \multicolumn{3}{c|}{\textbf{Depth}}                                                                                 &                                   &                                                       &                                                      &                                                        &                                                      &                                                      &                                                      \\ \cmidrule{3-5}
\multirow{-2}{*}{\textbf{Width}}     & \multirow{-2}{*}{\textbf{Block}}                                                & \textbf{Encoder}                               & \textbf{Middle}                    & \textbf{Decoder}                               & \multirow{-2}{*}{\textbf{Skip Connection}} & \multirow{-2}{*}{\textbf{PSNR}~$\uparrow$}                               & \multirow{-2}{*}{\textbf{SSIM}~$\uparrow$}                              & \multirow{-2}{*}{\textbf{FID}~$\downarrow$}                               & \multirow{-2}{*}{\textbf{LPIPS}~$\downarrow$}                               & \multirow{-2}{*}{\textbf{Params}~(M)~$\downarrow$}                         & \multirow{-2}{*}{\textbf{MACs}~(G)~$\downarrow$}                           \\ \midrule
\cellcolor[HTML]{FFC000}16  & TCFBlock [Ours]                                                         & {[}1,1,1,1{]}                         & 1                         & {[}1,1,1,1{]}                         & Sum                               & \cellcolor[HTML]{FACBCE}{\color[HTML]{333333} 17.789} & \cellcolor[HTML]{FBEAED}{\color[HTML]{333333} 0.620} & \cellcolor[HTML]{FCEDEF}{\color[HTML]{333333} 104.107} & \cellcolor[HTML]{FBD0D3}{\color[HTML]{333333} 0.359} & \cellcolor[HTML]{63BE7B}{\color[HTML]{333333} 1.67}  & \cellcolor[HTML]{63BE7B}{\color[HTML]{333333} 3.11}   \\
\cellcolor[HTML]{FFC000}32  & TCFBlock [Ours]                                                         & {[}1,1,1,1{]}                         & 1                         & {[}1,1,1,1{]}                         & Sum                               & \cellcolor[HTML]{FBEDF0}{\color[HTML]{333333} 18.706} & \cellcolor[HTML]{FBFAFD}{\color[HTML]{333333} 0.654} & \cellcolor[HTML]{DEF0E5}{\color[HTML]{333333} 85.439}  & \cellcolor[HTML]{FCF5F8}{\color[HTML]{333333} 0.305} & \cellcolor[HTML]{7EC992}{\color[HTML]{333333} 6.01}  & \cellcolor[HTML]{7FC993}{\color[HTML]{333333} 11.79}  \\
\cellcolor[HTML]{FFC000}64  & \cellcolor[HTML]{9BC2E6}TCFBlock [Ours]                                 & \cellcolor[HTML]{ACB9CA}{[}1,1,1,1{]} & \cellcolor[HTML]{FFD966}1 & \cellcolor[HTML]{ACB9CA}{[}1,1,1,1{]} & \cellcolor[HTML]{AEAAAA}Sum       & \cellcolor[HTML]{DAEFE2}19.150                        & \cellcolor[HTML]{93D2A4}0.671                        & \cellcolor[HTML]{C3E5CE}83.162                         & \cellcolor[HTML]{E3F1E9}0.291                        & \cellcolor[HTML]{EAF4EF}22.91                        & \cellcolor[HTML]{EEF6F2}45.86                         \\
\cellcolor[HTML]{FFC000}96  & TCFBlock [Ours]                                                         & {[}1,1,1,1{]}                         & 1                         & {[}1,1,1,1{]}                         & Sum                               & \cellcolor[HTML]{BFE3CA}19.194                        & \cellcolor[HTML]{FBFBFE}0.657                        & \cellcolor[HTML]{D7EDDF}84.796                         & \cellcolor[HTML]{AADAB8}0.283                        & \cellcolor[HTML]{FBC3C5}50.80                        & \cellcolor[HTML]{FBC2C5}102.21                        \\
\cellcolor[HTML]{FFC000}128 & TCFBlock [Ours]                                                         & {[}1,1,1,1{]}                         & 1                         & {[}1,1,1,1{]}                         & Sum                               & \cellcolor[HTML]{63BE7B}{\color[HTML]{333333} 19.338} & \cellcolor[HTML]{FBF9FC}{\color[HTML]{333333} 0.652} & \cellcolor[HTML]{E6F3EC}{\color[HTML]{333333} 86.075}  & \cellcolor[HTML]{86CC99}{\color[HTML]{333333} 0.278} & \cellcolor[HTML]{F8696B}{\color[HTML]{333333} 89.67} & \cellcolor[HTML]{F8696B}{\color[HTML]{333333} 180.85} \\
64                          & \cellcolor[HTML]{9BC2E6}ConvNeXt~\cite{convnext} & {[}1,1,1,1{]}                         & 1                         & {[}1,1,1,1{]}                         & Sum                               & \cellcolor[HTML]{FAD4D7}{\color[HTML]{333333} 18.042} & \cellcolor[HTML]{FBF8FB}{\color[HTML]{333333} 0.650} & \cellcolor[HTML]{FCE3E6}{\color[HTML]{333333} 113.985} & \cellcolor[HTML]{FCE7EA}{\color[HTML]{333333} 0.325} & \cellcolor[HTML]{FBFBFE}{\color[HTML]{333333} 25.61} & \cellcolor[HTML]{FCFBFE}{\color[HTML]{333333} 51.18}  \\
64                          & \cellcolor[HTML]{9BC2E6}ResNet~\cite{palette}  & {[}1,1,1,1{]}                         & 1                         & {[}1,1,1,1{]}                         & Sum                               & \cellcolor[HTML]{FBF7FA}{\color[HTML]{333333} 18.966} & \cellcolor[HTML]{C9E8D3}{\color[HTML]{333333} 0.664} & \cellcolor[HTML]{FCFAFD}{\color[HTML]{333333} 90.542}  & \cellcolor[HTML]{FCF9FC}{\color[HTML]{333333} 0.300} & \cellcolor[HTML]{FBD5D8}{\color[HTML]{333333} 42.90} & \cellcolor[HTML]{FCE1E4}{\color[HTML]{333333} 74.56}  \\
64                          & \cellcolor[HTML]{9BC2E6}NAFBlock~\cite{nafnet}   & {[}1,1,1,1{]}                         & 1                         & {[}1,1,1,1{]}                         & Sum                               & \cellcolor[HTML]{FBF9FC}19.042                        & \cellcolor[HTML]{93D2A4}0.671                        & \cellcolor[HTML]{CAE7D3}83.689                         & \cellcolor[HTML]{FCFCFF}0.295                        & \cellcolor[HTML]{EFF7F4}23.78                        & \cellcolor[HTML]{EEF6F3}45.87                         \\
64                          & TCFBlock [Ours]                                                         & {[}1,1,1,1{]}                         & \cellcolor[HTML]{FFD966}2 & {[}1,1,1,1{]}                         & Sum                               & \cellcolor[HTML]{B8E1C4}{\color[HTML]{333333} 19.205} & \cellcolor[HTML]{9AD5AB}{\color[HTML]{333333} 0.670} & \cellcolor[HTML]{EDF6F2}{\color[HTML]{333333} 86.717}  & \cellcolor[HTML]{EAF4EF}{\color[HTML]{333333} 0.292} & \cellcolor[HTML]{FCF3F5}{\color[HTML]{333333} 30.02} & \cellcolor[HTML]{F3F8F7}{\color[HTML]{333333} 47.48}  \\
64                          & TCFBlock [Ours]                                                         & {[}1,1,1,1{]}                         & \cellcolor[HTML]{FFD966}3 & {[}1,1,1,1{]}                         & Sum                               & \cellcolor[HTML]{6EC385}{\color[HTML]{333333} 19.321} & \cellcolor[HTML]{93D2A4}{\color[HTML]{333333} 0.671} & \cellcolor[HTML]{FCFBFE}{\color[HTML]{333333} 89.161}  & \cellcolor[HTML]{F8FAFB}{\color[HTML]{333333} 0.294} & \cellcolor[HTML]{FCE2E5}{\color[HTML]{333333} 37.13} & \cellcolor[HTML]{F8FAFC}{\color[HTML]{333333} 49.09}  \\
64                          & TCFBlock [Ours]                                                         & \cellcolor[HTML]{ACB9CA}{[}1,1,3,1{]} & 1                         & \cellcolor[HTML]{ACB9CA}{[}1,1,1,1{]} & Sum                               & \cellcolor[HTML]{85CC98}{\color[HTML]{333333} 19.285} & \cellcolor[HTML]{F9FBFC}{\color[HTML]{333333} 0.658} & \cellcolor[HTML]{FCEFF2}{\color[HTML]{333333} 101.486} & \cellcolor[HTML]{DBEFE3}{\color[HTML]{333333} 0.290} & \cellcolor[HTML]{F6F9F9}{\color[HTML]{333333} 24.78} & \cellcolor[HTML]{FCF3F6}{\color[HTML]{333333} 58.90}  \\
64                          & TCFBlock [Ours]                                                         & \cellcolor[HTML]{ACB9CA}{[}1,1,3,1{]} & 1                         & \cellcolor[HTML]{ACB9CA}{[}1,3,1,1{]} & Sum                               & \cellcolor[HTML]{8CCF9E}{\color[HTML]{333333} 19.274} & \cellcolor[HTML]{63BE7B}{\color[HTML]{333333} 0.677} & \cellcolor[HTML]{63BE7B}{\color[HTML]{333333} 74.908}  & \cellcolor[HTML]{63BE7B}{\color[HTML]{333333} 0.273} & \cellcolor[HTML]{FCFCFF}{\color[HTML]{333333} 25.78} & \cellcolor[HTML]{FCEFF2}{\color[HTML]{333333} 62.16}  \\
64                          & TCFBlock [Ours]                                                         & {[}1,1,1,1{]}                         & 1                         & {[}1,1,1,1{]}                         & \cellcolor[HTML]{AEAAAA}Concat~\cite{unet}    & \cellcolor[HTML]{F8696B}{\color[HTML]{333333} 15.134} & \cellcolor[HTML]{F8696B}{\color[HTML]{333333} 0.337} & \cellcolor[HTML]{F8696B}{\color[HTML]{333333} 236.843} & \cellcolor[HTML]{F8696B}{\color[HTML]{333333} 0.507} & \cellcolor[HTML]{EEF6F3}{\color[HTML]{333333} 23.61} & \cellcolor[HTML]{F5F9F9}{\color[HTML]{333333} 48.03}  \\
64                          & TCFBlock [Ours]                                                         & {[}1,1,1,1{]}                         & 1                         & {[}1,1,1,1{]}                         & \cellcolor[HTML]{AEAAAA}Attention~\cite{cotattention} & \cellcolor[HTML]{FAD3D6}{\color[HTML]{333333} 18.017} & \cellcolor[HTML]{FBEAED}{\color[HTML]{333333} 0.619} & \cellcolor[HTML]{FCDBDE}{\color[HTML]{333333} 121.777} & \cellcolor[HTML]{FBD3D6}{\color[HTML]{333333} 0.354} & \cellcolor[HTML]{FCFCFF}{\color[HTML]{333333} 25.97} & \cellcolor[HTML]{FCF7FA}{\color[HTML]{333333} 55.36} \\
\bottomrule
\end{tabular}}
\end{table*}

\begin{table*}[t]
\centering
\caption{Quantitative comparison of cloud removal performance and efficiency between DiffCR and existing models on two datasets.}
\label{tab:main}
\setlength{\tabcolsep}{2pt}
\begin{tabular}{l|cccc|cccc|cc}
\toprule
\multirow{2}{*}{\textbf{Method}} & \multicolumn{4}{c|}{\textbf{Sen2\_MTC\_Old}} & \multicolumn{4}{c|}{\textbf{Sen2\_MTC\_New}} & \multirow{2}{*}{\textbf{Params~(M)~$\downarrow$}} & \multirow{2}{*}{\textbf{MACs~(G)~$\downarrow$}} \\
\cmidrule(r){2-5} \cmidrule(r){6-9} & PSNR~$\uparrow$                   & SSIM~$\uparrow$ & FID~$\downarrow$ & LPIPS~$\downarrow$                 & PSNR~$\uparrow$              & SSIM~$\uparrow$ & FID~$\downarrow$ & LPIPS~$\downarrow$            &                         &                       \\
\midrule
MCGAN~\cite{mcgan} (CVPRW 2017)                     & 21.146  & 0.481  & 166.804 & 0.477 & 17.448          & 0.513          & 147.057         & 0.447          & 4.42                     & 71.56                      \\
Pix2Pix~\cite{pix2pix} (CVPR 2017)                   & 22.894  & 0.437  & 223.446 & 0.557 & 16.985          & 0.455          & 164.524         & 0.535          & 11.41                    & 58.94                      \\
AE~\cite{ae} (ECTI-CON 2018)                        & 23.957  & 0.800  & 169.347 & 0.439 & 15.100          & 0.441          & 206.134         & 0.602          & 6.53                     & \textbf{35.72}                      \\
STNet~\cite{stnet} (TGRS 2020)                     & 26.321  & 0.834  & 146.057 & 0.438 & 16.206          & 0.427          & 161.683         & 0.503          & 4.64                     & 304.31                     \\
DSen2-CR~\cite{dsen2-cr} (ISPRS J PHOTOGRAM 2020)                     & 26.967  & 0.855  & 123.382 & 0.330 & 16.827          & 0.534          & 140.208         & 0.446         & 18.92                   & 1240.23                    \\
STGAN~\cite{stgan} (WACV 2020)                     & 26.186  & 0.734  & 150.562 & 0.388 & 18.152          & 0.587          & 182.150         & 0.513          & 231.93                   & 1094.94                    \\
CTGAN~\cite{ctgan} (ICIP 2022)                     & 26.264  & 0.808  & 192.270  & 0.472 & 18.308          & 0.609          & 128.704         & 0.384          & 642.92                   & 632.05                     \\
CR-TS-Net~\cite{cr-ts-net} (TGRS 2022)                     & 26.900  & 0.857  & 121.447  & 0.325 & 18.585          & 0.615          & 96.364         & 0.342          & 38.68                   & 7602.97                     \\
PMAA~\cite{pmaa} (ECAI 2023)                      & 27.377  & 0.861  & 120.393 & 0.367 & 18.369          & 0.614          & 118.214         & 0.392          & 3.45                    & 92.34                      \\
UnCRtainTS~\cite{uncrtaints} (CVPRW 2023)                      & 26.417  & 0.837  &  130.875 & 0.400 & 18.770          & 0.631          & 93.509         & 0.333          & 0.56                     & 37.16 \\
\midrule
DDPM-CR~\cite{ddpm-cr}~(1000 steps)  & 27.060  & 0.854  & 110.919 &  0.320 & 18.742          & 0.614          &  94.401         & 0.329          & 445.44                     & 852.37 \\
\rowcolor[RGB]{217,217,217} 
\textbf{DiffCR [Ours] (only 1 step)}             & \textbf{29.112}       & \textbf{0.886}      & \textbf{89.845}       & \textbf{0.258}     & \textbf{19.150} & \textbf{0.671} & \textbf{83.162} & \textbf{0.291} & 22.91                    & 45.86\\                  
\bottomrule
\end{tabular}
\end{table*}

\subsubsection{Architecture Changes in Denoising Model}\label{sec:architecture}

Since the introduction of the diffusion models~\cite{thermodynamics,ddpm}, most studies have focused on improving their internal diffusion mechanisms to achieve high-quality samples~\cite{ddim,guided-diffusion} or fast inference speed\cite{pvdm,dpm-solver}. However, little research has been done to explore what kind of denoising neural model is more effective in the diffusion framework. Motivated by this, we propose a denoising model baseline with a UNet\cite{unet}-like architecture. It consists of several TCFBlocks that we designed. Through comprehensive ablation experiments, we explore the impact of architectural changes in the denoising model on cloud removal performance, with quantitative results shown in Table~\ref{tab:architecture}. Specifically, we explore the following architectural changes: 
\begin{itemize}
    \item The width (channels) of the denoising model.
    \item The block types of the denoising model.
    \item The depth of stacking multiple blocks.
    \item The fusion types of features from skip connections.
\end{itemize}
As shown in Table~\ref{tab:architecture}, we find that an increase in depth and width can significantly improve cloud removal performance. Among various block types (such as ResNet~\cite{resnet}, ConvNeXt~\cite{convnext} and NAFBlock~\cite{nafnet}), our proposed TCFBlock achieves the best performance, and adding more blocks in any part of the encoder, middle layer, or decoder leads to some performance improvement. Notably, the performance improvement is most significant when the encoder and decoder employ a symmetric structure in terms of depth (as in the third row from the bottom). Moreover, the most straightforward skip connection (Sum, i.e., directly adding encoder and decoder features of the same resolution) achieves the best performance, while more complex skip connections (such as Concat~\cite{unet}, which concatenates features in the channel dimension) lead to training instability, possibly due to the increased learning complexity. Selectively connecting features from the encoder and decoder using attention mechanisms~\cite{cotattention} can compromise performance, potentially due to loss of information resulting from such connections, affecting the quality of the generated images. Finally, for a balance between performance and efficiency, we opt for the data represented by the third row of the table as the architecture configuration for the denoising model, as it outperforms all existing methods with lower numbers of parameters and computational complexity. 

\subsection{Comparison with State-of-the-Art Methods}

\begin{figure*}[!t]
    \begin{minipage}[t]{0.5\linewidth}
        \centering
        \includegraphics[width=\textwidth]{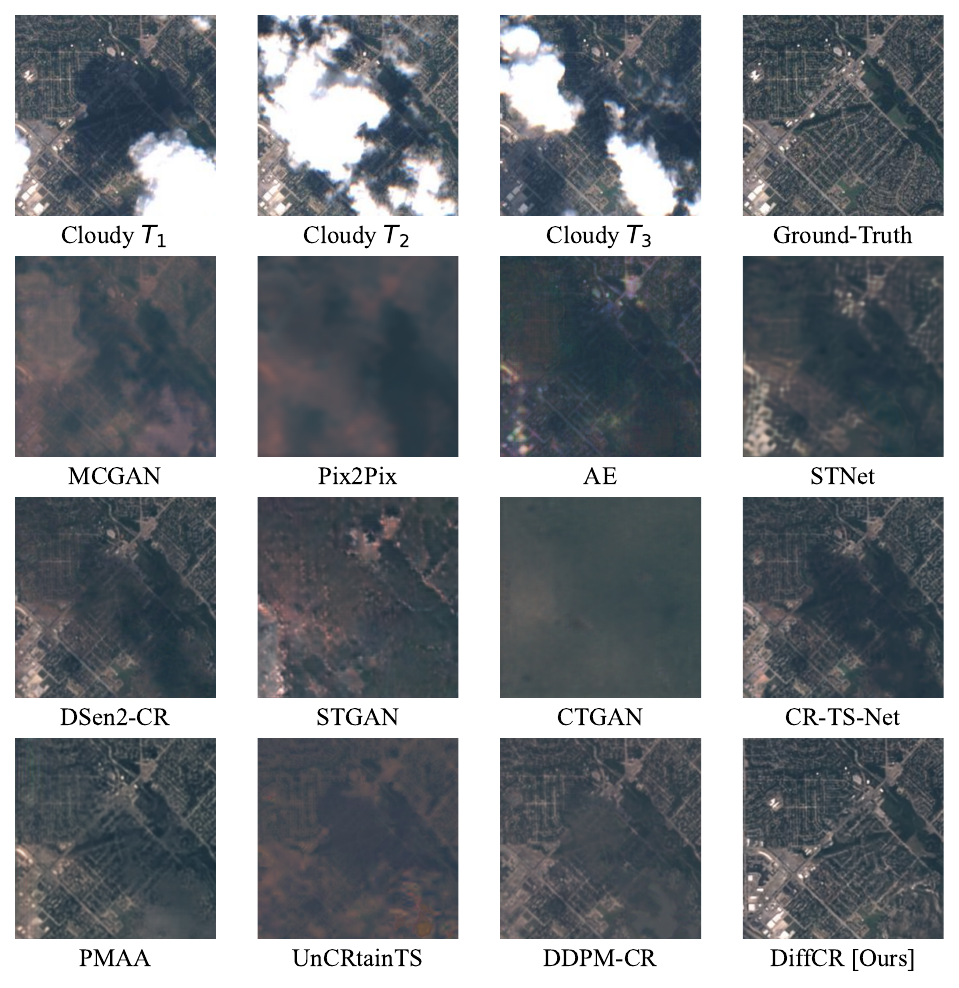}
        \centerline{(a)}
    \end{minipage}%
    \hspace{0.5em}\rule{0.5pt}{9.5cm}\hspace{0.5em}
    \begin{minipage}[t]{0.5\linewidth}
        \centering
        \includegraphics[width=\textwidth]{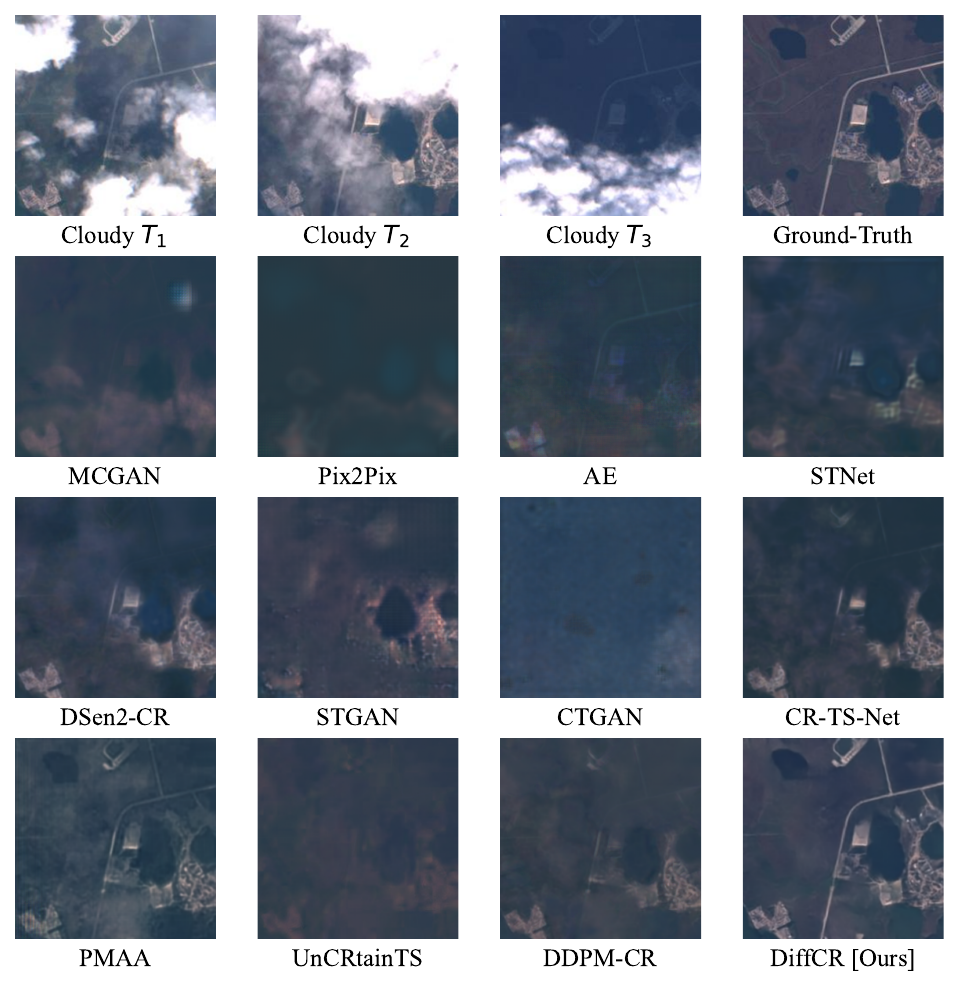}
        \centerline{(b)}
    \end{minipage}
    \caption{Qualitative comparison of the visualization results on the Sen2\_MTC\_Old dataset between our proposed DiffCR and other existing methods.}
    \label{fig:vis_old}
\end{figure*}

\begin{figure*}[!t]
    \begin{minipage}[t]{0.5\linewidth}
        \centering
        \includegraphics[width=\textwidth]{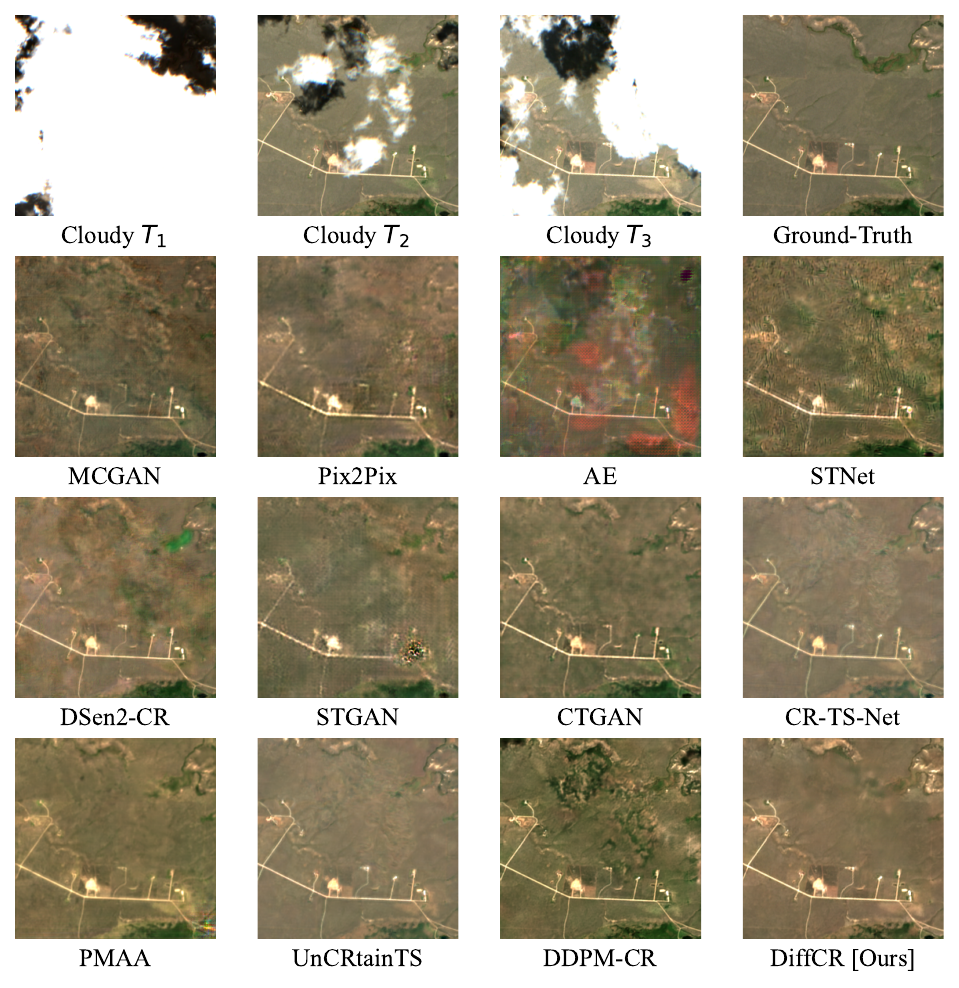}
        \centerline{(a)}
    \end{minipage}%
    \hspace{0.5em}\rule{0.5pt}{9.5cm}\hspace{0.5em}
    \begin{minipage}[t]{0.5\linewidth}
        \centering
        \includegraphics[width=\textwidth]{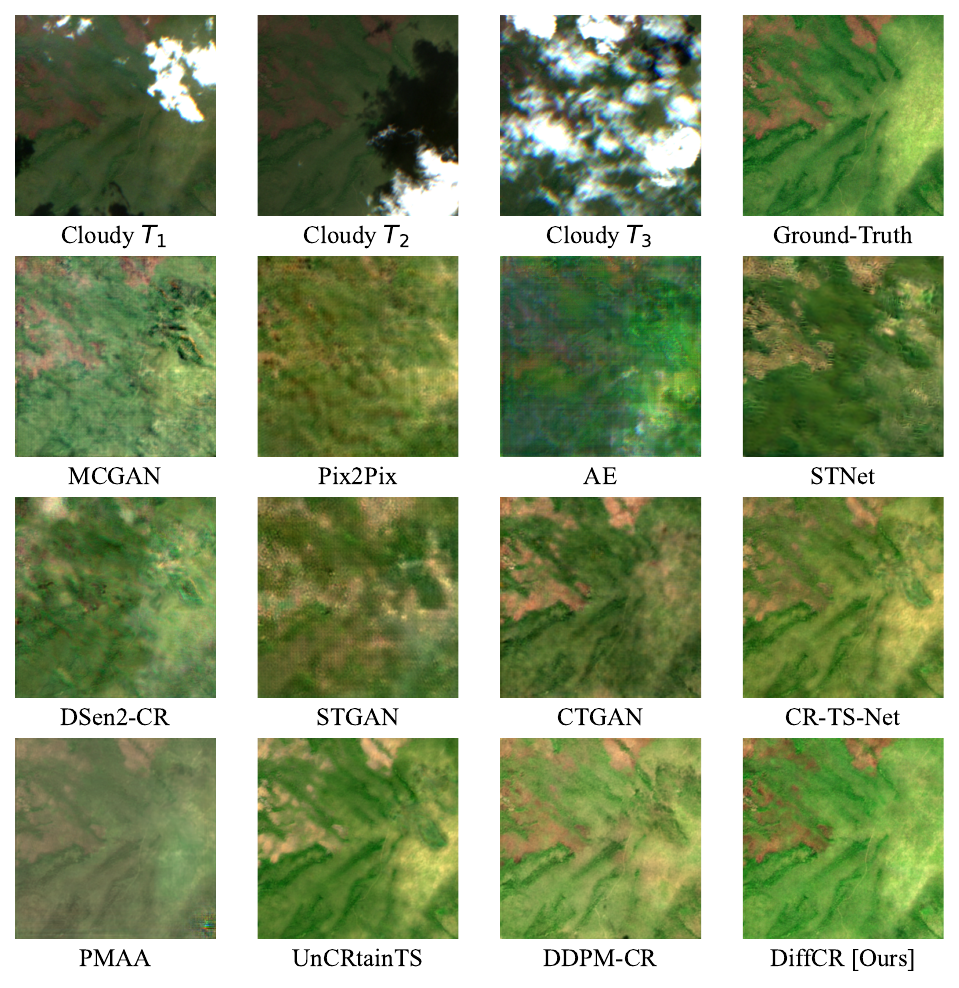}
        \centerline{(b)}
    \end{minipage}
    \caption{Qualitative comparison of the visualization results on the Sen2\_MTC\_New dataset between our proposed DiffCR and other existing methods.}
    \label{fig:vis_new}
\end{figure*}

\noindent{Table~\ref{tab:main} presents a quantitative comparison of cloud removal performance and efficiency between our proposed DiffCR and existing models, namely MCGAN~\cite{mcgan}, Pix2Pix~\cite{pix2pix}, AE~\cite{ae}, STNet~\cite{stnet}, DSen2-CR~\cite{dsen2-cr}, STGAN~\cite{stgan}, CTGAN~\cite{ctgan}, CR-TS-Net~\cite{cr-ts-net}, PMAA~\cite{pmaa}, UnCRtainTS~\cite{uncrtaints}, and DDPM-CR~\cite{ddpm-cr}. Except for DDPM-CR and DiffCR, which are based on diffusion methods, all others are based on either GAN or regression (without adversarial loss). The term ``1 step" refers to the sampling step used during the reverse inference phase.}

\subsubsection{Performance of Cloud Removal}

On two benchmark datasets, DiffCR has achieved consistent SOTA performance on all evaluation metrics of cloud removal, demonstrating the full potential of our proposed method. Specifically, DiffCR achieves an SSIM of 0.886 and 0.671 and an LPIPS of 0.258 and 0.291 on Sen2\_MTC\_Old and Sen2\_MTC\_New datasets, respectively. In contrast, existing models exhibit much lower performance values. Overall, the diffusion-based cloud removal approach, as exemplified by the proposed DiffCR model, outperforms all existing cloud removal models based on GAN or regression techniques. Compared to the existing diffusion-based model DDPM-CR, DiffCR achieves significant performance improvements on both datasets. Specifically, on the Sen2\_MTC\_Old dataset, DiffCR achieves a PSNR of 29.112, which is 2.052 higher than DDPM-CR's PSNR of 27.060, and an LPIPS of 0.258, which is 0.062 lower than DDPM-CR's LPIPS of 0.320. On the Sen2\_MTC\_New dataset, DiffCR achieves a PSNR of 19.150, which is 0.408 higher than DDPM-CR's PSNR of 18.742, and an LPIPS of 0.291, which is 0.038 lower than DDPM-CR's LPIPS of 0.329. These results demonstrate that DiffCR is superior to the existing diffusion-based model DDPM-CR in terms of accuracy and efficiency.

\subsubsection{Efficiency of Cloud Removal}

It’s worth noting that the above superior results are attained while maintaining a reasonable balance between the model’s parameters (22.91M) and the MACs (45.86G), which is a crucial aspect of the practical application of the model. Compared with models based on GAN or regression, DiffCR surpasses most models in terms of parameter count and computational complexity. Compared with the previous SOTA diffusion-based model DDPM-CR, DiffCR achieves higher performance with only 5.1\% and 5.4\% of the parameter count and computational complexity, respectively. Therefore, our work allows diffusion models to be applied in practical cloud removal tasks for the first time, laying the foundation for future research in this area.

As shown in Fig.~\ref{fig:vis_old} and Fig.~\ref{fig:vis_new}, we also provide a qualitative comparison between DiffCR and all other existing methods. All samples were randomly selected from the test set. The first row shows the input cloudy image (sequence length of $N=3$), with the last column representing the reference cloud-free image. The remaining rows display the reconstruction results of different algorithms. It can be observed that DiffCR outperforms all other methods in both local structure generation (as shown in subfigure a) and global brightness restoration (as shown in subfigure b), achieving the highest image fidelity.

Overall, these experimental results demonstrate the superior performance of DiffCR over existing SOTA methods, particularly in terms of image fidelity and computational efficiency. This underscores the effectiveness of our proposed approach, which utilizes data prediction and a conditionally decoupled diffusion model for cloud removal in optical satellite images.

\subsection{Failure Cases and Limitations}

\begin{figure}[!t]
\centering
\includegraphics[width=\linewidth]{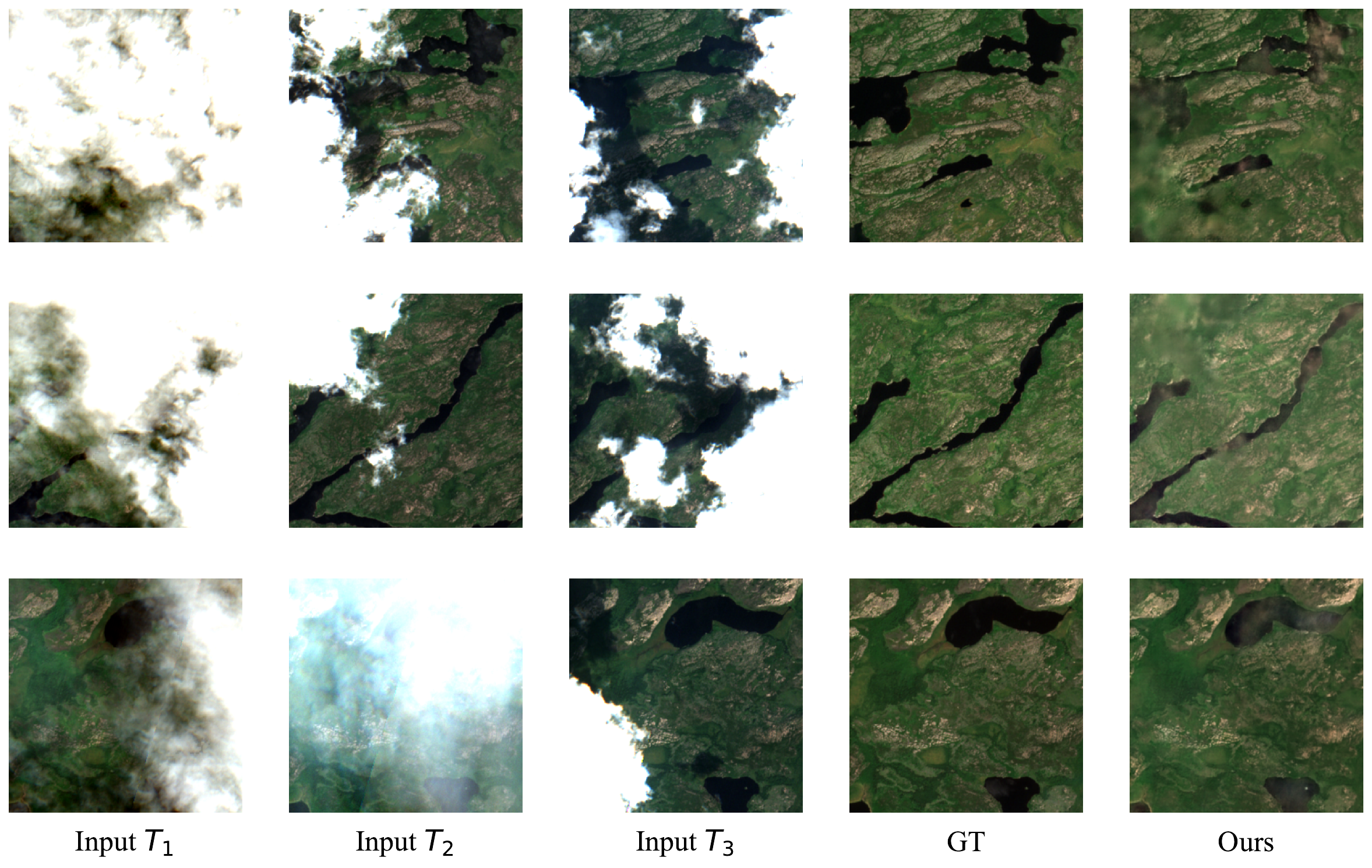}
\caption{Three failure examples. DiffCR exhibits inadequacy in scenarios where the input cloudy images feature lake formations resembling black holes. In such cases, the clouds may erroneously occupy the holes in the generation.}
\label{fig:failure}
\end{figure}


\noindent{Despite the exemplary performance demonstrated by our method, it still exhibits certain limitations. Firstly, when the input cloudy images containing lake regions with black hole-like shapes, our method encounters challenges in accurately reconstructing the lake areas (see Fig.~\ref{fig:failure}). This is attributed to the inherent similarity between the irregular, dark lake regions and cloud shadows. Secondly, although our method strives to strike a balance between performance and efficiency by employing lightweight and efficient modules, the inherent mechanisms of the diffusion model necessitate a considerable number of iterative steps during the model training phase in order to achieve convergence. Lastly, while our method is designed specifically for the task of cloud removal and has not undergone comprehensive validation across multiple tasks, its inherent generality enables its potential application in diverse domains, such as research pertaining to generative artificial intelligence from text to image and from image to image synthesis.}

\section{Conclusion}

\noindent{In this work, we propose DiffCR, a fast conditional diffusion framework for high-quality cloud removal. Existing cloud removal models based on GANs and regression fail to generate high-fidelity images, while diffusion-based models require thousands of iterations to ensure high sampling quality. Thus, balancing sampling speed and sample quality has become the main challenge in cloud removal. In contrast, DiffCR redefines the conditional diffusion framework for cloud removal, designs a novel conditionally decoupled neural model, and parameterizes the denoising model by directly predicting clean data. We also develop a novel and efficient time and condition fusion block within the model. As a result, DiffCR generates high-fidelity and detail-rich cloud-free images in just one sampling step. Through comprehensive experiments on two benchmark datasets, we demonstrate that DiffCR outperforms the best publicly available models regarding synthesis quality.}





{
\bibliographystyle{IEEEtran}
\bibliography{refs}
}

\begin{IEEEbiography}[{\includegraphics[width=1in,height=1.25in, clip,keepaspectratio]{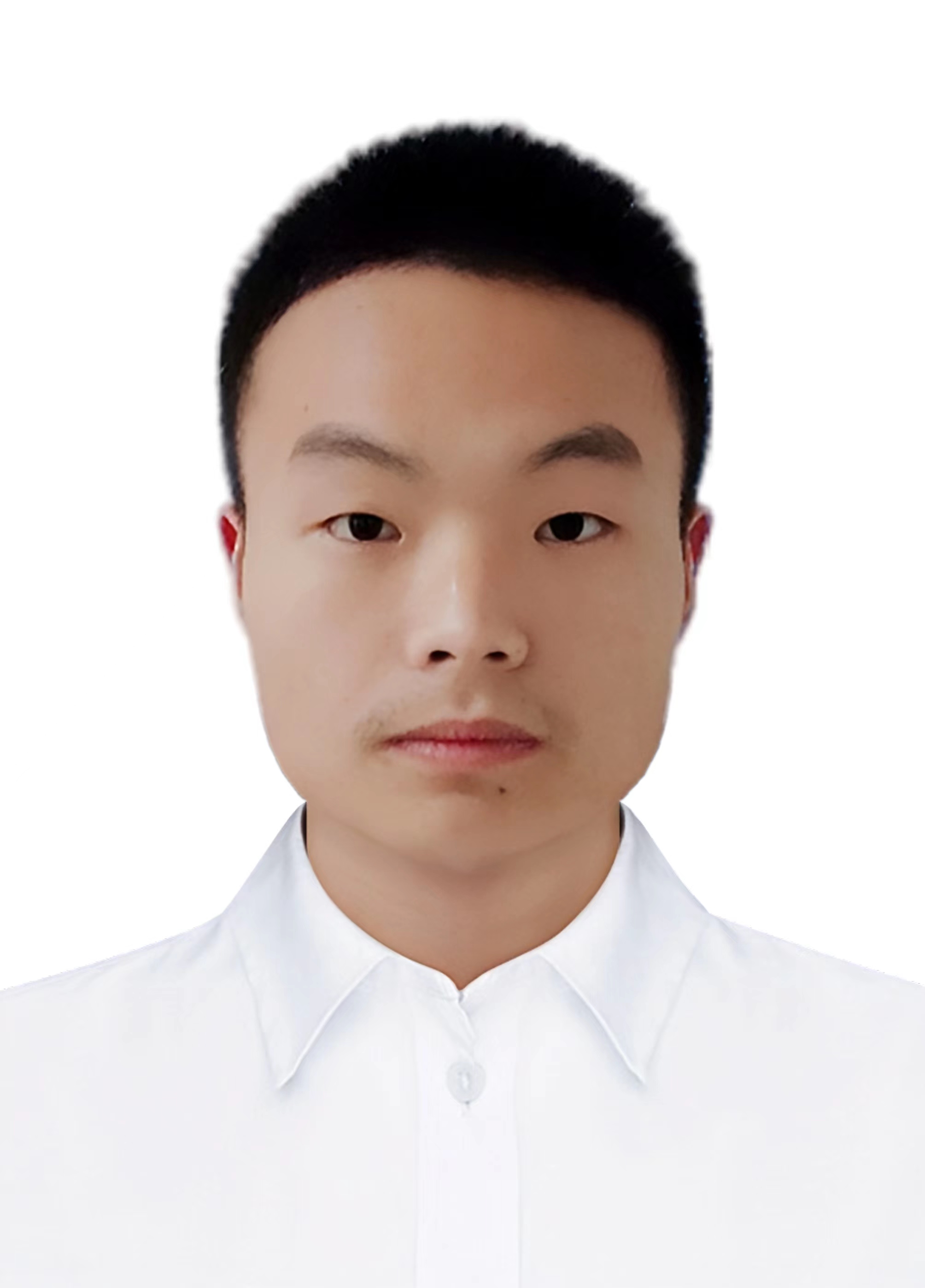}}]{Xuechao Zou} received the B.E. degree from Qinghai University, Xining, China, in 2021. He is pursuing an M.S. degree with the Department of Computer Technology and Application at Qinghai University. His research interests primarily focus on computer vision, particularly in the area of remote sensing image processing.
\end{IEEEbiography}

\begin{IEEEbiography}[{\includegraphics[width=1in,height=1.25in, clip,keepaspectratio]{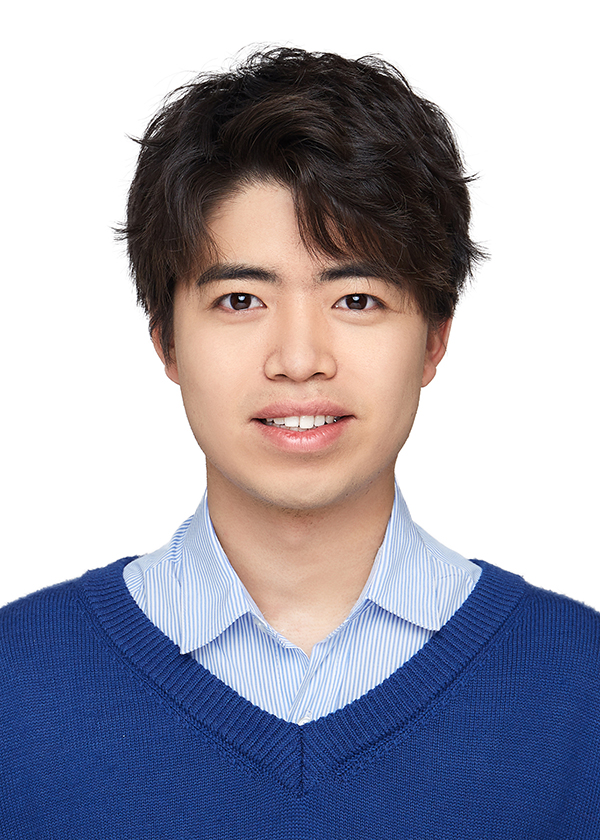}}]{Kai Li} received the B.E. degree from Qinghai University, China, in 2020. He is pursuing an M.S. degree with the Department of Computer Science and Technology at Tsinghua University, China. His research interests are primarily in audio processing and audio-visual learning, especially speech separation.
\end{IEEEbiography}
\vspace{-8mm}

\begin{IEEEbiography}[{\includegraphics[width=1in,height=1.25in, clip,keepaspectratio]{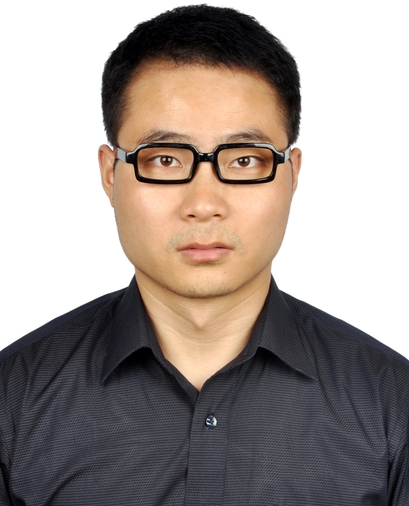}}]{Junliang Xing} received his dual B.S. degrees in computer science and mathematics from Xi’an Jiaotong University, Shaanxi, China, in 2007, and the Ph.D. degree in Computer Science and Technology from Tsinghua University, Beijing, China, in 2012. He is currently a Professor at the Department of Computer Science and Technology, Tsinghua University, Beijing, China. Dr. Xing has published over 100 peer-reviewed papers with more than 12,000 citations from Google Scholar. His current research interest mainly focuses on computer gaming problems related to single/multiple agent learning and human-computer interactive learning.
\end{IEEEbiography}
\vspace{-8mm}

\begin{IEEEbiography}[{\includegraphics[width=1in,height=1.25in, clip,keepaspectratio]{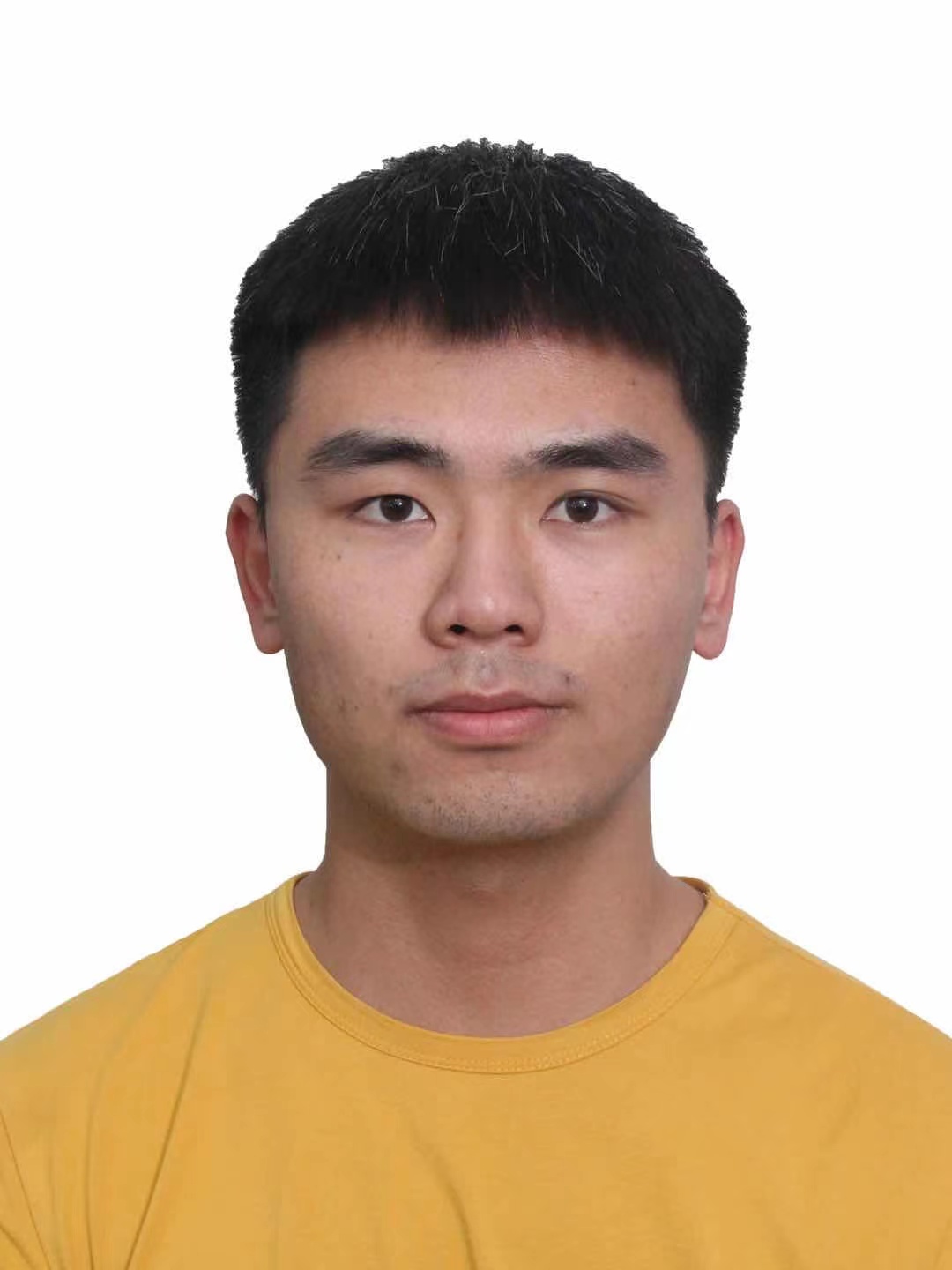}}]{Yu Zhang} received the B.E. degree from Jilin Jianzhu University, Changchun, China, in 2021. He is pursuing an M.S. degree with the Department of Computer Technology and Application at Qinghai University. His research interests include deep single and multi-agent reinforcement learning.
\end{IEEEbiography}
\vspace{-8mm}

\begin{IEEEbiography}[{\includegraphics[width=1in,height=1.25in, clip,keepaspectratio]{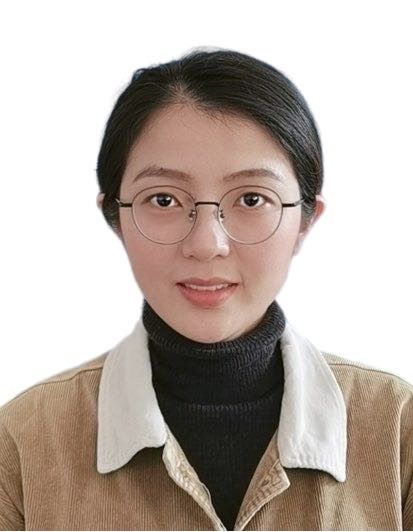}}]{Shiying Wang} received the B.E. degree from Jilin Agricultural University, Changchun, China, in 2015. She is pursuing a Ph.D. degree, at Qinghai University. Her research interests include remote sensing image fusion, especially pan-sharpening.
\end{IEEEbiography}
\vspace{-8mm}

\begin{IEEEbiography}[{\includegraphics[width=1in,height=1.25in, clip,keepaspectratio]{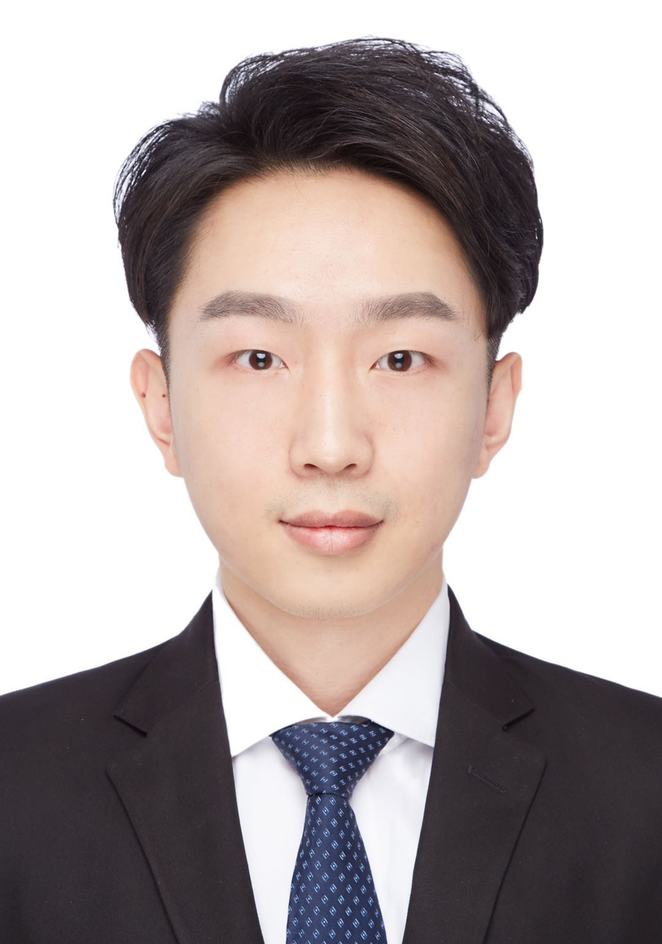}}]{Lei Jin} is currently an associate research fellow in the Beijing University of Posts and Telecommunications (BUPT), Beijing, China. Before that, he graduated from the same university with a Ph.D. degree in 2020. He received a bachelor's degree in the BUPT in 2015. His research interest includes machine learning and pattern recognition, with a focus on 6Dof pose estimation and Human pose estimation.  What’s more, he concentrated on Network security and analyzed traffic security during pursuing his Ph.D. degree.
\end{IEEEbiography}
\vspace{-8mm}

\begin{IEEEbiography}[{\includegraphics[width=1in,height=1.25in, clip,keepaspectratio]{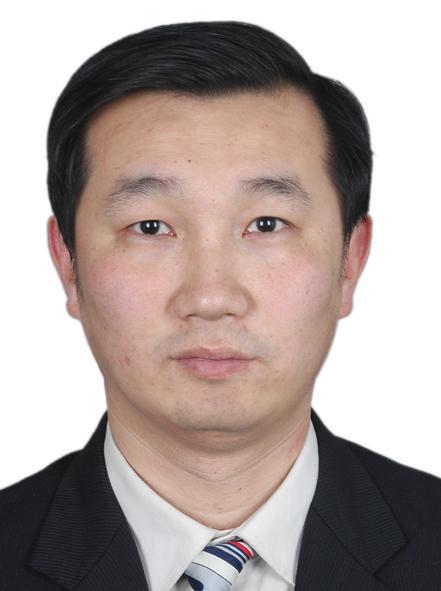}}]{Pin Tao} received his B.S. degree and Ph.D. in computer science and technology from Tsinghua University, Beijing, China, in 1997 and 2002. He is
currently an Associate Professor at the Department of Computer Science and Technology, Tsinghua University, Beijing, China. Dr. Tao has published more than 80 papers and over 10 patents. His current research interests mainly focus on human-AI hybrid intelligence and multimedia-embedded processing.
\end{IEEEbiography}

\vfill

\end{document}